\documentclass{article} 
\usepackage{iclr2025_conference,times}

\usepackage{amsmath,amsfonts,bm}

\def\eqref#1{equation~\ref{#1}}

\def\1{\bm{1}}

\DeclareMathAlphabet{\mathsfit}{\encodingdefault}{\sfdefault}{m}{sl}
\SetMathAlphabet{\mathsfit}{bold}{\encodingdefault}{\sfdefault}{bx}{n}

\usepackage{hyperref}
\usepackage{url}
\usepackage{graphicx}
\usepackage{booktabs}
\title{Shortcut Before Circuit: Document Statistics Time In-Context Conflict Resolution}

\author{Yijun Liao \\
\texttt{liuyingliao0620@gmail.com} \\
  \And
  Fanwei Liang \\
  \texttt{202466331178@mail.scut.edu.cn}
}

\newcommand{\budget}{16000}        

\newif\ifarxiv
\arxivtrue          

\iclrfinalcopy 
\begin{document}

\maketitle

\newcommand{\todo}[1]{\textcolor{red}{#1}}

\begin{abstract}
When a context asserts two values for one fact, a model commits to a cue---recency,
repetition, position---but natural data rarely makes these disagree, so behavior
cannot reveal which. We train 26M-parameter transformers on a synthetic language
where recency and rarity are exactly coextensive, and separate them with a minimal
causal edit that inverts one cue while holding the truth, token count and answer
position fixed. All 75 runs reach accuracy $\geq 0.999$, including where the trivial
heuristic fails, so no held-in evaluation distinguishes them. Under intervention the
per-cell readout does not replicate: 13 of 25 cells differ by more than 0.3 in sign
fraction across three seeds, the largest by 0.879 against a standard error of 0.025. The
construction predicts this---coextensive rules leave the objective indifferent between
them---and the variance is ordered by how much of the optimization each comparison
releases. What replicates is timing: escape from a positional shortcut with a closed-form
ceiling, monotone in redundancy. Probed before that escape, attribution reverses sign
in 32 of 75 runs at unchanged accuracy, and gating on circuit formation is necessary
but not sufficient. The corpus fixes when a mechanism appears, not which one --- a criterion for when
mechanistic attribution to data is available at all, and our construction makes the
unavailable case exact.
\end{abstract}

\section{Introduction}

A document may assert that a slot holds one value and later assert that it holds
another. Answering a query about that slot requires committing to a cue, and in
ordinary text the cues agree: the value stated later is usually also the value
stated once rather than many times. Because they agree, accuracy reveals nothing
about which cue a model uses. The commitment surfaces only when they come
apart---a stale retrieval corpus, a partially edited document, an inconsistently
propagated update---and by then the choice was fixed during pretraining.

Documents in our synthetic assignment language are sequences of (entity,
attribute, value) statements followed by a query; the answer is the most recent
value assigned to the queried slot. Bindings are resampled per document and the
training stream is infinite, so the task cannot be solved from weights. We vary
two statistics of the cue structure while holding the task, its Bayes-optimal
policy, and the entire training recipe fixed: $R_{\mathrm{old}}$, the number of
times a superseded value is repeated, and $\Delta D$, the number of statements
from the last update to the query. Because the current value appears exactly
once and each superseded value appears $R_{\mathrm{old}}$ times, ``take the
rarest value'' and ``take the most recent value'' select the same value on every
training document, so no observational analysis can separate them. We therefore
read out the rule with a minimal causal edit that inverts the multiplicities of
the two competing values, leaving the ground truth, the token count, the answer
position, and every other token unchanged. The signed change in log-odds,
$\Delta$, is positive when the readout tracks rarity and negative when
multiplicity boosts the value the model should suppress.

Every analyzed cell reaches $\mathrm{acc} \geq 0.999$, including on the stratum where
the trivial ``the last update is the answer'' heuristic fails, so two mechanisms with
opposite dependence on repetition count receive the same score from any held-in
evaluation. Under intervention the same models separate: pooled over three seeds the
sign fraction rises monotonically by redundancy level through $R_{\mathrm{old}}=12$, while a
control edit on a non-queried slot perturbs the same number of tokens and stays within
0.015 nats. Before acquiring retrieval, models converge on a rule that copies the value
at a fixed token offset from the end, whose ceiling
$1/\lvert\mathrm{supp}(\Delta D)\rvert$ follows from the token layout and which they
saturate to within 2\% at narrow support. Applying our probe during this phase yields
strongly negative $\Delta$, which read as a rule attribution would place a clean
frequency-type region at low $R_{\mathrm{old}}$ --- produced by models with no
retrieval circuit at all (\S\ref{sec:plateau}).

Taken together these give a criterion rather than a map. A mechanism can be
attributed to a corpus statistic when the objective is not indifferent between the
alternatives, and cannot when it is. On the indifferent side our construction makes
the indifference exact, and we show the consequence three ways: the per-cell readout
moves 0.879 across seeds where documents give a standard error of 0.025, its
within-run standard deviation across post-formation checkpoints reaches 0.303, and
doubling the schedule moves cells that no data statistic distinguishes in advance. On the determined side timing
survives every one of those comparisons. What the corpus fixes is when a mechanism
appears, not which one.

\section{Related Work}

Behavioral studies of knowledge conflict establish that models resolve conflicting
evidence systematically, with preferences tracking entity popularity and evidence
coherence \citep{longpre2021entity,xie2024adaptive}, update plausibility
\citep{kortukov2024studying}, the number of agreeing sources
\citep{chen2022rich,jin2024tug}, surface repetition \citep{yan2024understanding} and
source authority \citep{schuster2026whose}, and over-attention to superseded state
in multi-session dialogue \citep{liao2025inertia}. These measure which cue wins in a deployed
model but cannot attribute the preference to data: the corpus is fixed and, for the
cues at issue, unobservable, and because natural documents rarely make the candidate
cues disagree, any observed preference is confounded with whatever cue happened to
correlate with the answer. Mechanistic work localizes the components carrying
conflicting information \citep{jin2024cutting}, with end-to-end circuits demonstrated for non-trivial behaviors in trained models,
though completeness remains difficult to certify \citep{wang2022ioi}, and identifies prefix-matching
induction heads as a canonical copy-based mechanism associated with in-context
learning \citep{olsson2022incontext}, traced in detail \citep{singh2024needs} and shown
to be non-monotone \citep{singh2023transient}, while controlled-corpus studies establish
distributional properties governing whether in-context learning appears at all
\citep{chan2022data,raventos2023pretraining,bratulic2025balancing,kim2025training,han2023understanding},
with a parallel line on how parametric knowledge is stored and extracted
\citep{allenzhu2024physics}.
\citet{reddy2024mechanistic} is the closest antecedent for our timing result, tracing
the abrupt appearance of an induction head to the data distribution and identifying
progress measures that precede in-context learning, as our loss-derivative peak does.
What is explained there is \emph{when} a single mechanism appears, extended by
\citet{gibson2026distinct} to four algorithmic phases implemented by two qualitatively
distinct multi-layer subcircuits whose boundaries are set by data diversity; ours is
\emph{which} of two a model commits to once it has, with the appearance left intact in
every cell. In both lines the contrast is whether in-context retrieval is used, or
context against weights; the cues \emph{within} the context are never separated,
because natural data does not separate them.

Concurrent work establishes that mechanistic explanations are frequently non-unique.
Enumerating candidate explanations on Boolean functions and small networks yields
systematic non-identifiability: several circuits replicate a behavior, and one circuit
admits several interpretations \citep{meloux2025everything}. Structurally near-disjoint
circuits can be simultaneously faithful, sparse and complete on the same task
\citep{chen2026rome}, and circuit-level conclusions can fail to survive defensible
variation in the analysis pipeline itself \citep{mahale2026explanation}. These results
document multiplicity; none of them says when an explanation \emph{is} determined. Our
construction supplies one such condition on the data side: where two rules are
coextensive on the training distribution the objective cannot select between them, and
where they are not it can. The distinction is testable here because both sides are
measured in the same grid --- escape timing replicates across seeds, the rule readout
does not. The variance we report is across training runs rather than across analytic
choices, which is the axis those three studies hold fixed.

Our positional rule is a shortcut \citep{geirhos2020shortcut,mccoy2019right,liu2023lost}, unusual
in that the token layout is ours to fix, so its payoff follows in closed form along a
controlled axis rather than being diagnosed after the fact. Abrupt reorganizations are
often summarized as phase diagrams
\citep{power2022grokking,nanda2023progress,doshi2024grok,goddard2025when}, and claims
of abruptness require care since discontinuous metrics can manufacture transitions
from smooth change \citep{schaeffer2023mirage}; we report the transition in
unthresholded metrics alongside the gate. \citet{park2024competition} is closest, a
phase diagram for a synthetic in-context task with phases framed as competition
between algorithms, but their axes are properties of the task distribution and phases
are assigned by matching outputs against candidate algorithms. Ours are statistics of
the cue structure within a document, with the task and its Bayes-optimal policy
identical in every cell, so the in-distribution optimum is attained everywhere and
behavioral matching cannot distinguish the phases at all.

\section{Setup}
\label{sec:setup}

\subsection{A language with an aliased rule pair}
\label{sec:alias}

Documents are sequences of assignment statements followed by a query. Each
statement binds a value to an (entity, attribute) pair, which we call a
\emph{slot}, and is emitted as four tokens \texttt{[ent] [attr] [val] [SEP]};
the query appends \texttt{[QUERY] [ent] [attr] [ARROW]} and the target is the
value currently bound to the queried slot. Identifiers come from disjoint
vocabularies of size 200, 8 and 512. Bindings are resampled per document and the
stream is infinite, so the task is not solvable from weights.

Every document contains one queried slot: a value $v_{\mathrm{old}}$ is bound to
it and repeated $R_{\mathrm{old}}$ times, a later statement rebinds the slot to
$v_{\mathrm{new}}$, and the query follows $\Delta D$ statements after that
rebinding. Remaining positions are filled by statements from other slots, each
independently receiving an update with probability $p_{\mathrm{update}} = 0.5$.
Values are drawn without replacement within a document, so each identifier
belongs to at most one slot; this makes ``the answer value occurs exactly once
before the query'' structural rather than probabilistic, removing a baseline
noise term from every measurement.

\paragraph{The alias.} Because each superseded value is repeated
$R_{\mathrm{old}}$ times while the current value is written once, for the
queried slot $s$ of any document $d$
\begin{equation}
\underbrace{\arg\max_i \mathrm{pos}(s_i)}_{\textsc{recency}}
\;=\;
\underbrace{\arg\min_v \bigl\lvert \{\, i : \mathrm{val}(s_i) = v \,\} \bigr\rvert}_{\textsc{rarity}}
\qquad \text{for all } d,
\label{eq:alias}
\end{equation}
with ties in \textsc{rarity} broken toward the more recent occurrence.
Equation~\ref{eq:alias} is an identity of the generator, not a statistical
tendency: the two rules never disagree on a single training document, the
objective is indifferent between them, and no observational analysis can
separate them. Five further candidates are separable observationally; per-cell rates and
the closed forms they match are in App.~\ref{app:cells}.

\subsection{The two axes}
\label{sec:axes}

Neither axis changes the target function; both change how expensive each
mechanism is to implement.

\emph{Redundancy} $R_{\mathrm{old}} \in \{3,5,8,12,16\}$, fixed per configuration, sets
how often the language demonstrates that a value appearing more than once is not the
current one. Two lower values are excluded: at $R_{\mathrm{old}}=1$ \textsc{rarity}
carries no information and the edit has no domain, and at $R_{\mathrm{old}}=2$ the retrieval circuit fails to form within our budget in four of five cells
(App.~\ref{app:posneg}). The upper end is structural: superseded copies
are placed in a window of width $\mathrm{spread} \times n_{\mathrm{stmts}}$ before the
rebinding, and once $2R_{\mathrm{old}}$ approaches that width the window degenerates into a
solid block abutting the rebinding, making the queried slot identifiable without reading
the query. At $\mathrm{spread}=0.8$ and $n_{\mathrm{stmts}} \geq 45$ the largest
admissible value is 16.

\emph{Update-to-query distance} $\Delta D$ is the number of statements between
the final rebinding and the query. A recency mechanism must match the queried
slot and select its latest occurrence, attending across the $\Delta D$
intervening statements; suppressing repeated values within a slot has no such
dependence. We sample $\Delta D \sim U[\ell(d), h(d)]$ with
$\ell(d) = \max(1, \mathrm{round}(d/2))$ and
$h(d) = \max(\ell{+}1, \mathrm{round}(3d/2))$ for nominal
$d \in \{2,3,5,8,16\}$.

The support is an interval rather than a point because at constant $\Delta D$ the
answer sits at a fixed token offset from the end and ``copy the value at offset $k$''
becomes fully correct with no slot matching. Support width bounds such a rule at
$1/\lvert\mathrm{supp}(\Delta D)\rvert$, falling from 0.333 at $d \in \{2,3\}$ to
0.059 at $d=16$; \S\ref{sec:ceiling} derives this and shows models occupy it for
thousands of steps.

\subsection{Invariants and training}
\label{sec:inv}

Because the argument rests on the two axes being the only systematic difference
between cells, we verify five generator invariants on 1500 documents per cell before
training; four hold exactly and the fifth partially, with antecedent distance
exceeding the filler mean by 2--22\% (App.~\ref{app:gen}). Document length is sampled from $U[45,55]$ statements
independently of both axes, with the lower bound set by the most demanding cell
($n_{\mathrm{stmts}} \geq R_{\mathrm{old}} + \Delta D_{\max} + 5$) and then used
everywhere, since a cell-dependent length distribution would itself be a
covariate. We deliberately do not equalize the tail-update fraction
$(1-\rho)^{\Delta D}$: forcing it constant requires inserting distractor updates
in the tail, which at small $\Delta D$ creates a different perfect shortcut, so
we report it as a covariate and stratify held-in evaluation on it.

We train decoder-only transformers with 8 layers, $d_{\mathrm{model}}=512$,
8 heads, rotary embeddings and pre-norm blocks (26.1M parameters). Attention
uses per-head RMS normalization on queries and keys with a learnable per-dimension gain initialized to the constant 2.0 rather than 1.0; optimization moves it down from there rather than up (App.~\ref{app:qk}). Each configuration is trained for 16000 steps at batch size 256 with AdamW ($\mathrm{lr}=10^{-3}$, cosine
decay, weight decay 0.1, none on embeddings) in bfloat16, with loss on all
tokens and documents generated on the fly. The budget is set by the slowest cell:
the circuit forms before step 1000 at
$R_{\mathrm{old}}=16$ but near step 6000 at $R_{\mathrm{old}}=3$ (\S\ref{sec:dyn}), so
16000 steps leaves every analyzed cell at least 6000 post-formation steps, and all but two at least 9000. App.~\ref{app:drift} shows the readout is stable across an
8$\times$ budget range in the interior of the grid.

\section{Reading out the rule}
\label{sec:probe}

\subsection{The multiplicity inversion}

To separate aliased rules we apply minimal edits to held-out documents. The logic is
that of an interchange intervention \citep{geiger2021causal,geiger2024finding}, applied
at the input rather than to an internal variable. An edit
targets one rule, moves that rule's prediction, and leaves the ground truth
unchanged. Writing
$m(x; v^\star) = \log p(v^\star \mid x) - \log p(v_{\mathrm{truth}} \mid x)$,
the readout is the paired difference
\begin{equation}
\Delta = m(x_{\mathrm{edit}}; v^\star) - m(x_{\mathrm{base}}; v^\star),
\label{eq:readout}
\end{equation}
where both terms use the same contrast value $v^\star$, chosen as the target
rule's post-edit prediction and required to differ from the ground truth on both
sides.

The edit separating \textsc{rarity} from \textsc{recency} inverts value
multiplicity inside the queried slot: one copy of $v_{\mathrm{old}}$ is retained
and every other copy is rewritten as a copy of $v_{\mathrm{new}}$, so post-edit
counts are $\{v_{\mathrm{old}}\!:\!1,\, v_{\mathrm{new}}\!:\!R\}$, exchanged
relative to the base. Under this edit \textsc{recency} still predicts
$v_{\mathrm{new}}$ (the last statement of the slot is untouched),
\textsc{frequency} predicts $v_{\mathrm{new}}$, \textsc{primacy} predicts
$v_{\mathrm{old}}$ on both sides and cancels in Equation~\ref{eq:readout}, and
only \textsc{rarity} flips. The retained copy is the earliest, which makes any first-occurrence positional account
cancel between the two terms (App.~\ref{app:gen}).

\subsection{What we report}
\label{sec:dv}

The primary dependent variable is the fraction of held-out documents on which
$\Delta$ has the expected sign, over 400 documents at the final checkpoint, with an
exact binomial test. We report the median and interquartile range alongside it but treat
the sign fraction as primary. Per-document $\Delta$ is heavy-tailed, so the mean is
unstable where the sign fraction is not, and the median is bounded from above by
contrast-pair mass in exactly the high-effect cells where the separation is largest
(\S\ref{sec:degrade}); the sign fraction depends on neither magnitude nor scale, so it
compares across seeds and budgets, where the unit of replication is the run rather than
the document. Table~\ref{tab:stats} reports every summary we computed for seed 0,
Table~\ref{tab:spread} the sign fraction for all three.

\section{Results}
\label{sec:results}

\subsection{Behavior saturates everywhere}
\label{sec:sat}

Every analyzed cell reaches in-distribution accuracy $\geq 0.999$, including on the
stratum where no update follows the queried rebinding, so any evaluation restricted to
the training distribution assigns these 75 models the same score. This holds of the
analyzed grid, not the language: in the excluded $R_{\mathrm{old}}=2$ column three of
five cells terminate at another mechanism's analytic ceiling (App.~\ref{app:posneg}).
Behavioral saturation is a regime the data statistics must permit, not a given.

\subsection{The mechanism is ordered by redundancy}
\label{sec:flip}

\begin{figure}[ht]
\centering
\includegraphics[width=0.85\textwidth]{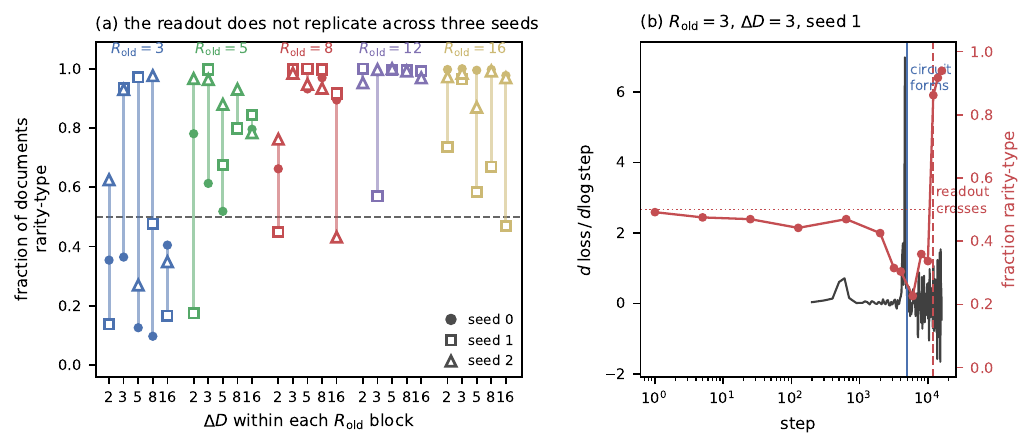}
\caption{\textbf{(a)} Sign fraction at 16000 steps for the 25 cells under three seeds,
joined per cell; filled circles seed 0, open squares seed 1, open triangles seed 2. Line
length is the range across seeds, reaching 0.879 against a binomial standard error of
0.025. \textbf{(b)} The $R_{\mathrm{old}}=3$, $\Delta D=3$ run of seed 1
(\S\ref{sec:plateau}): $d\,\mathrm{loss}/d\log\mathrm{step}$ (left axis) and sign
fraction (right axis). The interior peak marks circuit formation near step 4600; the
readout crosses near step 12000 at unchanged accuracy (\S\ref{sec:plateau}).}
\label{fig:phase}
\end{figure}

Under intervention the same models separate (Figure~\ref{fig:phase}), and pooled over
three seeds the sign fraction is ordered by redundancy level: row means run 0.479,
0.768, 0.858, 0.962 and 0.879 at $R_{\mathrm{old}}=3,5,8,12,16$, monotone through $12$
and breaking at $16$. Two of the three seeds are monotone through $12$ individually;
seed 2 falls from 0.905 at $5$ to 0.812 at $8$, so only the $3\to5$ and $8\to12$ steps
hold in every seed. The break at $16$ is one seed's: the row mean falls 0.225 in seed 1 and 0.024 in seed 2
while seed 0 does not fall at all, so nine tenths of the pooled decrease comes from
seed 1, whose three lowest cells there are among the widest-spread in
Table~\ref{tab:spread}. The exception is an instance of the non-replication rather than
a feature of the axis.

Per cell it does not reproduce, and we therefore do not assign a mechanism to a cell.
Thirteen of 25 cells span more than 0.3 across seeds, and in eight the three seeds do
not agree on which side of indifference the cell lies. The widest is
$R_{\mathrm{old}}=3$, $\Delta D=8$ at 0.098, 0.477 and 0.977, a range of 0.879,
whose endpoints are separately decisive at
$p = 1\times10^{-54}$ and $p = 5\times10^{-88}$ in opposite directions: either run alone
would license a confident attribution, and no number of documents within one run detects
the other. The spread is not a measurement artifact: within a single run the sign
fraction has median standard deviation 0.069 across post-formation checkpoints and the
range in the widest cells does not narrow over further training (App.~\ref{app:drift}),
while training is bit-identical on rerun at fixed seed (App.~\ref{app:gen}). We treat a cell-to-cell
difference as meaningful only when it exceeds the range in
Table~\ref{tab:spread}, which leaves the row ordering and rules out the per-cell map.

The low end is where the readout is least determined rather than where it reverses.
All five $R_{\mathrm{old}}=3$ cells of seed 0 have negative gated medians and a row mean
sign fraction of 0.270; the same row reads 0.538 and 0.630 in seeds 1 and 2, spanning
0.098 to 0.977 across the fifteen runs. It is also where the circuit forms most slowly
(\S\ref{sec:dyn}). The $\Delta D$ axis carries no ordering we interpret.

\subsection{What the sign rules out, and where}
\label{sec:sign}

The multiplicity inversion raises the occurrence count of the correct answer from
one to $R$ and lowers the superseded value to one, so any mechanism that accumulates
evidence with occurrence count predicts $\Delta < 0$: a context-frequency prior, or an
induction head aggregating over matches.
Pooled by row the gated median is positive at every $R_{\mathrm{old}} \geq 5$ in all
three seeds; per cell it fails in four, all within 0.031 nats of zero
($-0.030$, $-0.012$, $-0.009$, $-0.006$).
At the high end the readout is therefore inconsistent with accumulation and consistent
with suppression, opposing both the majority-rule preference reported under knowledge
conflict \citep{jin2024tug} and the repetition-driven token co-occurrence reinforcement
reported for in-context learning \citep{yan2024understanding} --- not tightly, since
those settings repeat across documents rather than within a slot. What the sign separates is a family
of mechanisms rather than our two named rules, at the level of the row.

\subsection{Covariates}
\label{sec:covar}

Three quantities shift with the axes and cannot be held fixed; we report them per cell
(Table~\ref{tab:covar}) rather than adjusting for them. Slot count falls with
$R_{\mathrm{old}}$; realized redundancy falls with $\Delta D$ because copies outside the
document canvas are discarded; and edit domain tracks realized redundancy, so the cells
with the weakest readout condition on the smallest subset.

The last admits a within-cell test, since initialization and data order are fixed within
a run and the spread of Table~\ref{tab:spread} therefore cannot enter it. Realized
redundancy is recorded per document, so we
split each cell at its median and ask whether the readout tracks it inside the cell as
across cells. It does, in whichever direction that cell points:
at $R_{\mathrm{old}}=16$, $\Delta D=2$ the high-redundancy half reads $+2.274$
against $+1.351$ (Table~\ref{tab:within}), and at $R_{\mathrm{old}}=3$,
$\Delta D=5$ --- where the seed-0
median is negative --- $-0.120$ against $-0.069$, more negative rather than less.
The gradient holds in 65 of the 75 runs, and six of the ten exceptions have
pooled medians below 0.15 nats in absolute value, where there is no effect to
stratify. This does not sit in tension with \S\ref{sec:flip}. The gradient runs in the direction of
each cell's own effect, so the corpus statistic sets how much the readout responds while
the run sets which way it points: sensitivity replicates, baseline does not. A single
seed would have reported both as one number. This is strong evidence
available to us that $R_{\mathrm{old}}$ acts through the statistic rather than through
document selection.

\subsection{Where the measurement degrades}
\label{sec:degrade}

Contrast-pair mass falls as the effect grows, reaching 0.59 in the worst cell of the
three seeds, which bounds the instrument. Gating per document at mass $\geq 0.5$
shifts the cell median by 2\% where mass is 0.98 and by 14--63\% in the ten cells
where mass falls below 0.85. Cells whose readout exceeds roughly 2 nats therefore cannot
be separated from one another by median; every median we quote in the main text for
such a cell is gated, and cross-cell differences in median are lower bounds, compressed
most where the effect is largest. The sign fraction is computed over all documents with
edit domain and is unaffected by the gate by construction (App.~\ref{app:stats}).

\subsection{What the objective determines, and what it does not}
\label{sec:flat}

The cross-seed spread of \S\ref{sec:flip} is not a failure of the instrument but a
consequence of Equation~\ref{eq:alias}. Because \textsc{recency} and \textsc{rarity} select the same value on every training
document, a model implementing one and a model implementing the other incur identical loss
on every training document: the objective is indifferent between them, and nothing in the
data selects one. This is underspecification \citep{damour2022underspecification} localized to a
mechanism: many predictors with equivalent held-in performance, differing in which of
two rules they implement rather than in which features they use. That claim predicts
\emph{where the variance lives}: it should scale with how much of the optimization a
comparison releases, not with how many documents the probe sees. It does. Against a binomial
standard error of 0.025 across documents, the within-run term across post-formation
checkpoints reaches 0.303 over 69 runs (App.~\ref{app:drift}), doubling the cosine
period moves a cell by up to 0.55 (\S\ref{sec:limits}), and the seed term reaches
0.879 --- four orders of comparison, ordered by how much optimization each releases. A
cell whose entry is one run at one checkpoint under one schedule therefore reports a
draw from that distribution rather than a property of the data statistics.
Nothing in behavior distinguishes the endpoints: all 75 runs sit at $\geq 0.999$ on both
strata and on the copy diagnostic. The classification analogue is predictive
multiplicity \citep{marx2020multiplicity}.

Consistently, the loss derivative has exactly one interior peak in all 50 runs of seeds 0
and 1 and none where the readout changes sign (\S\ref{sec:plateau}), which excludes a
non-flat reorganization as the driver of the crossing.

We also tried to measure the geometry directly and report the attempt as uninformative:
the direction that trades the two rules is flat relative to the loss gradient, but so is
every direction we construct the same way from seven other readouts, including two the
objective constrains (App.~\ref{app:flat}). At $2.6\times10^{7}$ parameters the
measurement does not separate a coextensive pair from a separable one, so the argument
rests on Equation~\ref{eq:alias} and on the variance ordering above.

\section{Shortcut Dynamics}
\label{sec:dyn}

Training does not proceed monotonically toward retrieval, and the route differs by cell.
Where the circuit forms slowly and the shortcut pays well, models spend thousands of
steps on a plateau implementing a positional rule; where it forms quickly they pass
through a low-accuracy regime and acquire retrieval without ever saturating the
shortcut.

\subsection{The plateau has a closed-form ceiling and models saturate it}
\label{sec:ceiling}

Statements occupy exactly four tokens, so the answer's offset measured backward from
the end is $4\Delta D + 6$, independent of document length. A
rule that ignores the query and emits the value at a fixed offset is therefore correct
exactly when $\Delta D$ takes one particular value, so
\begin{equation}
\mathrm{posCeil} \;=\; \max_{d} \Pr[\Delta D = d] \;=\; \frac{1}{\lvert \mathrm{supp}(\Delta D) \rvert}
\label{eq:posceil}
\end{equation}
since $\Delta D$ is uniform on its support (App.~\ref{app:posceil} derives both, and
states the two conditions under which the bound could be exceeded).
Table~\ref{tab:ceiling} compares this to plateau accuracy in five cells: the two at
$|\text{supp}| = 3$ saturate the bound to within 2\% and agree with each other to
0.0002, while the wider-support cells attain 70--73\% of a ceiling that is itself
falling, with one exception that attains neither (App.~\ref{app:posceil}).

Saturating Eq.~\ref{eq:posceil} shows only that a positional rule suffices, not that
retrieval is absent. For that we use the copy diagnostic: accuracy on in-context value
tokens repeating an earlier value of the same slot, which needs slot matching but is
scored on tokens the positional rule cannot reach. Throughout the plateau it reads
0.001--0.046 against sequence accuracies of 0.077--0.326 in the four cells that reach a
plateau (Table~\ref{tab:ceiling}).

\subsection{Escape time is set jointly by both statistics}
\label{sec:escape}

Escape is abrupt on the evaluation grid: the copy diagnostic moves from near
chance to above 0.95 between consecutive evaluations in every cell that escapes,
and sequence accuracy jumps from Eq.~\ref{eq:posceil} to 1.000. Because a discrete gate
on a coarse grid can manufacture a transition from smooth change
\citep{schaeffer2023mirage}, we locate the reorganization in the training loss
instead, which is recorded every 100 steps and involves no threshold. Its
derivative with respect to $\log$ step has exactly one interior peak in every one
of the 50 runs of seeds 0 and 1, a few hundred steps before the gate is crossed. The peak
locates a reorganization without certifying which one, which is why the gate is on the
copy diagnostic and not on the loss curve.

The peak also locates escape at 100-step rather than 1000-step resolution, which
resolves the grid floor. Row means of the peak position are 5500 and 6100 steps at
$R_{\mathrm{old}}=3$, 1740 and 2660 at $5$, 740 and 1180 at $8$, 400 and 420 at $12$,
and 400 and 400 at $16$, for seeds 0 and 1: escape is monotone in $R_{\mathrm{old}}$
across four separable levels in both seeds, where the gate separated only three. Seed 2
gives 4220, 1960 and 820 at $R_{\mathrm{old}}=3,5,8$, so the row means of the three seeds
do not overlap below $R_{\mathrm{old}}=12$: 4220--6100 steps at $3$ against 1740--2660 at
$5$ and 740--1180 at $8$.
$R_{\mathrm{old}}=12$ and $16$ are the pair the derivative does not separate, both
sitting within one recording interval of 400 steps, so the four levels are $3$, $5$, $8$
and $\{12, 16\}$ rather than five. Peak
height increases with escape step across the 50 runs of seeds 0 and 1 (Spearman
$\rho = 0.933$ on average ranks), from
1.60 at step 400 to 14.48 at 9600. Where escape is very early the peak is small, and
the cosine schedule's final decay exceeds it in three runs of seed 0
($16$, $\Delta D \in \{2,3,5\}$) and four of seed 2 ($12$, $\Delta D \in \{8,16\}$ and
$16$, $\Delta D \in \{5,16\}$), so at $R_{\mathrm{old}} \geq 12$ we rely on the gate.
Neither axis
predicts escape alone: at $16$ every cell of seeds 0 and 1 escapes by step 400 regardless
of band width, and at $3$ escape spans 1900--9600 steps across the three seeds, broadly
inverse with band width but not monotone in it. The ordering is consistent with a race between the circuit's
loss reduction and the shortcut's, which we state as a reading rather than an isolated
mechanism. Per-run values and both views of the peak: Table~\ref{tab:escape} and
Figure~\ref{fig:traj}, App.~\ref{app:states}.

\subsection{Attribution on the plateau reverses the conclusion}
\label{sec:plateau}

Applying the probe of \S\ref{sec:probe} before escape yields strongly negative
$\Delta$ throughout the main grid, not only in the excluded cells. On the last
probe point preceding escape, the sign fraction is below 0.20 in 32 of the 75 runs
and below 0.05 in 22 of them, spanning every row across the three seeds
($-3.36$ with 0.01 of documents positive at $R_{\mathrm{old}}=16$, $\Delta D=2$;
$-4.02$ with 0.01 at $12, 8$; $-1.32$ with 0.02 at $3, 5$). The excluded
$R_{\mathrm{old}}=2$ cells (App.~\ref{app:posneg}) are the terminal version of the
same phenomenon rather than a separate case.

Read as a rule attribution this is a clean ``frequency-type'' region: multiplicity
boosting the superseded value, exactly the pattern the majority-rule literature would
predict. It is an artifact --- the models producing it have no retrieval circuit, so
the reading cannot reflect a contest between recency and rarity whatever else it
reflects. A positional account of the sign is natural but not testable from data-side
quantities alone (App.~\ref{app:posneg}). The reading is not uniformly
negative: in five runs the pre-escape sign fraction is above 0.6, and in all five the
last pre-escape probe sits within one evaluation interval of escape, so we cannot
separate a genuinely positive plateau reading from a probe that caught the transition.

Nothing in task performance signals that the quantity being measured has changed
meaning: in the $R_{\mathrm{old}}=3$, $\Delta D=2$ run at the 32000-step schedule
$\Delta$ crosses from $-0.587$ to $+1.802$ across one evaluation interval with
accuracy 1.000 at both ends. We therefore gate every cell on the copy diagnostic
before reporting $\Delta$, classify each run as retrieval, positional, or neither, and
report the classification alongside the read-out (App.~\ref{app:states}). Read-outs
are taken at a fixed budget of \budget{} steps; post-escape trajectories drift in both
directions (App.~\ref{app:drift}).

\paragraph{Gating on formation is necessary but not sufficient.} In the
$R_{\mathrm{old}}=3$, $\Delta D=3$ cell of seed 1 the copy diagnostic reaches 1.000 at
step 5000 and the readout stays frequency-type for a further 7000 steps (sign fractions
0.23, 0.36, 0.34 at steps 6000, 8000, 10000) before crossing to 0.86 at 12000 and
settling at 0.94. Accuracy is 1.000 at every one of those points and the loss
derivative has no peak near step 12000, its four largest values falling at steps 4600,
4700, 4300 and 15400 (Figure~\ref{fig:phase}, right). A gate on circuit formation would
have admitted this run at step 6000 and returned the opposite conclusion --- the
readout moves 0.71 in sign fraction with no signature in behavior or in loss, which is
\S\ref{sec:flat} in a single run.

\section{Discussion}
\label{sec:disc}

The construction was built to ask which cue a model commits to and answers a different
question: when a cue can be attributed to a model at all. Where two rules are
coextensive on the training distribution the loss does not choose between them, so a
measurement of which one a model implements is a measurement of where that run stopped
along a direction the objective does not constrain. Such measurements are not
meaningless; their unit of replication is the run. The claim is conditional on coextensiveness rather than on scale. What a larger model or a
natural corpus would change is which pairs of rules are coextensive and therefore which
readouts are underdetermined, not whether coextensiveness has this consequence; the
synthetic language is what makes the condition exact enough to check.

Three practices follow for any phase diagram whose cells are mechanisms. Report seed
variance per cell: a per-document $p$-value speaks to sampling within one run, and none
of our conclusions would survive a protocol reporting one seed per cell
\citep{vanderwal2025polypythias}. Gate on circuit formation rather
than task accuracy, since 32 of our 75 runs return a confident reversed attribution
before the circuit exists, with no behavioral signal that the probe is measuring
nothing. And treat that gate as necessary but not sufficient: even a gated
single-checkpoint readout can report either direction.

What survives is what the objective is not indifferent to: escape timing, monotone in
redundancy across four levels in seeds 0 and 1 and three in seed 2, recovered from an
unthresholded quantity; the shortcut's closed-form ceiling, saturated to within 2\%; and the
within-cell dose-response, which holds in 65 of the 75 runs with the run fixed.
Timing is a property of the corpus here. Mechanism identity, at the resolution of one
cell, is not.

This argues for aliased constructions, not against them. The alias makes the
underdetermination visible; in natural data the same indifference exists wherever cues
agree, but there a single-seed measurement returns a confident number with nothing to
contradict it. What remains open is the comparison this construction cannot make:
every cell is aliased by Equation~\ref{eq:alias}, so we have no non-aliased control in
which to measure the same geometry and find it absent.

\section{Limitations}
\label{sec:limits}

\emph{The endpoint of the readout is set jointly by the corpus and the optimizer.}
Retraining a cell with an identical seed and data stream but a cosine schedule of twice
the length moves the terminal sign fraction by up to 0.55
(Table~\ref{tab:budget}). Three cells that escape at step 1000 under both budgets
isolate the schedule from the escape step: two move (0.45 to 0.98 and 0.47 to 0.79) and
one does not (0.58 to 0.54, the endpoint itself indistinguishable from indifference
at $p=0.11$, and flat across 18 post-formation checkpoints), while
saturated cells are stable. One is converging slowly toward a value it has not reached
and another is not converging, and nothing in the data distinguishes them in advance. We
therefore report the map at one budget and one schedule, and cannot say how much of the
cross-seed spread at 16000 steps is incomplete convergence rather than a different
endpoint. The question we set out to answer was which cue wins; the one we can answer is when
that question has an answer a corpus can supply.

\emph{Two collinearities, one partly broken.} The positional ceiling
$1/\lvert\mathrm{supp}(\Delta D)\rvert$ falls from 0.333 to 0.059 across the
$\Delta D$ axis, and the readout at $R_{\mathrm{old}}=16$ falls with it. A fixed-width
band ($\Delta D \sim U[d, d{+}8]$) holds the ceiling at 0.111; run at
$R_{\mathrm{old}}=3$ in three seeds it leaves the cross-seed spread intact
(App.~\ref{app:fixband}), so that spread is not mediated by shortcut payoff --- but the $R_{\mathrm{old}}=16$ gradient itself remains untested. Slot count falls with $R_{\mathrm{old}}$ and cannot be held fixed together with everything
else it moves. Matching it by shortening the documents recovers 85--97\% of the gap in sign
fraction across three seeds but only 17--74\% of the gap in median, with the residual
redundancy confound running counter to the observed change in all three
(App.~\ref{app:slotmatch}); matching it at fixed length by moving $p_{\mathrm{update}}$
instead removes almost none of the dispersion difference, and there the two cells do not converge under
one pooling statistic and converge almost completely under another (App.~\ref{app:pupd}).
The length--sparsity complex carries the direction at this point in the grid; which
component of it does is unresolved.

\emph{Two of our five data knobs admit no causal readout.} With explicit update
markers the edit changes sequence length; with history-indexed queries the target
rule is no longer the last value, so Equation~\ref{eq:readout} does not apply.
For these we can report observational agreement but not a mechanism, and we
exclude them rather than report a readout we cannot validate.

\emph{The queried slot is not perfectly indistinguishable.} Spreading $R_{\mathrm{old}}$
copies over a fixed window leaves the queried slot's copies more dispersed than a filler
slot's (App.~\ref{app:gen}), so a model could locate it without reading the query. Every
analyzed run reaches 1.000 on the copy diagnostic, which requires genuine slot matching,
so the cue is at most a redundant path --- but it is collinear with both axes and we
cannot rule out a contribution.

\emph{One architecture family, one task family.} The grid is 8 layers at
$d_{\mathrm{model}}=512$ (26.1M parameters) on a single synthetic language, with a depth
arm at 4 and 12 layers on the $\Delta D = 5$ column (App.~\ref{app:depth}). Escape
ordering and the within-cell gradient survive both depths, while the per-cell readout
moves as much with depth as with seed and in a seed-dependent direction, so depth is a
further draw from the direction of \S\ref{sec:flat} rather than a control on it. Depth and
capacity covary at fixed width; with no ordering in depth, there is nothing for that
separation to explain. The QK-normalization gain is a choice of the same kind: halving it
removes the circuit from the slowest-forming cell within our budget (App.~\ref{app:qk}),
so the analyzed grid is defined partly by it.

\subsubsection*{Reproducibility statement}

The language and its two axes are fully specified in \S\ref{sec:alias}--\S\ref{sec:axes},
training in \S\ref{sec:inv}, and the edit with its five invariants in
\S\ref{sec:probe}. Generator invariants (App.~\ref{app:gen}), the state
classification that gates every readout (App.~\ref{app:states}), and per-cell
collision rates (App.~\ref{app:cells}) are measured rather than asserted. \ifarxiv  Code
reproducing every table, together with the generator, the edit suite and the
per-cell readout logs, is at \url{https://github.com/lyj20071013/Shortcut-Before-Circuit}.
\else Anonymized code reproducing every table is included with the submission.\fi

\bibliography{iclr2025_conference}
\bibliographystyle{iclr2025_conference}

\appendix
\section{Generator measurements}
\label{app:genmeas}

Everything in this appendix is measured on 1500 documents per cell before training.
These are properties of the generator, not of any trained model.

\subsection{Invariants and failure modes}
\label{app:gen}

\begin{table}[h]
\centering
\small
\caption{Generator invariants, 1500 documents per cell. Columns are defined in the
text below: \emph{bindMax} cross-document binding reuse, $r_{\mathrm{len}}$
correlation between realized $\Delta D$ and token count, \emph{ansLast} trivial-copy
documents, \emph{updDens} pre-query update density in the final decile,
\emph{adj}/\emph{ant} same-slot adjacency rate and antecedent distance for the queried
($q$) and filler ($f$) slots. The fifth invariant holds only partially: ant$_q$ exceeds
ant$_f$ by 2--22\%.}
\label{tab:inv}
\begin{tabular}{rrrrrrrrrr}
\toprule
$R_{\mathrm{old}}$ & $\Delta D$ & bindMax & $r_{\mathrm{len}}$ & ansLast &
updDens & adj$_q$ & adj$_f$ & ant$_q$ & ant$_f$ \\
\midrule
3 & 2 & 3 & $-0.024$ & 0 & 1.029 & 0.056 & 0.078 & 13.32 & 10.92 \\
3 & 3 & 3 & $-0.023$ & 0 & 1.020 & 0.058 & 0.077 & 13.21 & 10.95 \\
3 & 5 & 3 & $+0.028$ & 0 & 1.022 & 0.056 & 0.074 & 13.09 & 11.00 \\
3 & 8 & 3 & $-0.035$ & 0 & 1.021 & 0.060 & 0.077 & 12.55 & 10.92 \\
3 & 16 & 3 & $-0.002$ & 0 & 0.984 & 0.074 & 0.076 & 11.49 & 11.03 \\
\midrule
5 & 2 & 3 & $-0.003$ & 0 & 1.074 & 0.087 & 0.104 & 9.57 & 8.29 \\
5 & 3 & 3 & $-0.003$ & 0 & 1.071 & 0.088 & 0.105 & 9.53 & 8.31 \\
5 & 5 & 3 & $+0.009$ & 0 & 1.013 & 0.090 & 0.107 & 9.30 & 8.34 \\
5 & 8 & 3 & $+0.007$ & 0 & 1.015 & 0.091 & 0.106 & 9.22 & 8.35 \\
5 & 16 & 4 & $-0.013$ & 0 & 1.008 & 0.092 & 0.106 & 8.54 & 8.36 \\
\midrule
8 & 2 & 3 & $+0.031$ & 0 & 1.041 & 0.135 & 0.152 & 6.93 & 6.30 \\
8 & 3 & 3 & $+0.031$ & 0 & 1.031 & 0.135 & 0.151 & 6.91 & 6.32 \\
8 & 5 & 3 & $-0.039$ & 0 & 0.994 & 0.135 & 0.150 & 6.88 & 6.34 \\
8 & 8 & 3 & $-0.003$ & 0 & 0.981 & 0.132 & 0.150 & 6.81 & 6.29 \\
8 & 16 & 3 & $-0.008$ & 0 & 1.006 & 0.138 & 0.155 & 6.52 & 6.22 \\
\midrule
12 & 2 & 3 & $+0.007$ & 0 & 1.102 & 0.180 & 0.197 & 5.35 & 5.00 \\
12 & 3 & 3 & $+0.007$ & 0 & 1.101 & 0.181 & 0.196 & 5.33 & 5.01 \\
12 & 5 & 3 & $-0.036$ & 0 & 1.083 & 0.183 & 0.198 & 5.36 & 5.00 \\
12 & 8 & 3 & $+0.006$ & 0 & 0.946 & 0.189 & 0.200 & 5.28 & 4.92 \\
12 & 16 & 3 & $-0.043$ & 0 & 1.056 & 0.188 & 0.206 & 5.16 & 4.84 \\
\midrule
16 & 2 & 3 & $+0.022$ & 0 & 1.065 & 0.213 & 0.236 & 4.71 & 4.22 \\
16 & 3 & 3 & $+0.022$ & 0 & 1.059 & 0.213 & 0.237 & 4.71 & 4.23 \\
16 & 5 & 2 & $+0.001$ & 0 & 1.097 & 0.212 & 0.249 & 4.74 & 4.16 \\
16 & 8 & 3 & $-0.018$ & 0 & 1.052 & 0.213 & 0.248 & 4.69 & 4.16 \\
16 & 16 & 2 & $-0.024$ & 0 & 1.048 & 0.218 & 0.260 & 4.54 & 4.02 \\
\bottomrule
\end{tabular}
\end{table}

\textbf{bindMax}: the largest number of distinct documents in which any single
(slot, value) triple appears. Bounded by 4 throughout, consistent with chance
collision at $200 \times 8 \times 512$ possible triples and 75000 statements
sampled, so bindings are not memorizable across documents. Note that counting
raw occurrences rather than documents would return $\geq R_{\mathrm{old}}$ by
construction, since the queried slot holds $R_{\mathrm{old}}$ copies of one
value within a single document.

\textbf{$r_{\mathrm{len}}$}: Pearson correlation between realized $\Delta D$ and
document token count, $|r| \leq 0.043$ in every cell. Without this the $\Delta D$
axis would confound distance with length.

\textbf{ansLast}: documents in which the answer value equals the last value token
before the query, which would admit a trivial copy rule. Zero in all 25 cells,
by construction rather than by chance --- values are drawn without replacement
within a document.

\textbf{updDens}: density of update statements in the final decile of the
pre-query region, divided by the global density, computed over filler slots only.
Within $[0.946, 1.102]$ throughout. The queried slot's own rebinding is excluded
because it sits at a fixed offset $m - \Delta D - 1$ and would land in the final
decile at small $\Delta D$, reporting the construction as a violation.

\textbf{adj, ant}: same-slot adjacency rate and mean distance to nearest
same-slot antecedent, for the queried slot ($q$) and filler slots ($f$).
Adjacency is matched within 0.042 in every cell. Antecedent distance is not:
ant$_q$ exceeds ant$_f$ by 2--22\%, largest at $R_{\mathrm{old}}=3$,
$\Delta D=2$ where $13.32 / 10.92 = 1.22$. This follows from spreading
$R_{\mathrm{old}}$ copies over a window of width
$\mathrm{spread} \times n_{\mathrm{stmts}} = 0.8 \times 50 = 40$ statements:
$40 / 3 = 13.3$, matching the measured value. The gap varies along both axes and
is discussed as a limitation in \S\ref{sec:limits}. Three earlier versions of the generator failed one of these invariants. The most
consequential had a bias that scaled in $R_{\mathrm{old}}$, collinear with our main
axis and therefore invisible to any cross-cell comparison.

\paragraph{A dataloader failure mode that manufactures a phase transition.} An
earlier version of the loader reused the same seed offset across passes, replaying a
fixed document set rather than drawing from the infinite stream. Training loss then
falls below the analytic entropy floor of the task while held-out loss rises, and the
crossing looks like an abrupt reorganization on the training curve. It is entirely
memorization, and it is worth stating because the diagnostic we rely on in
\S\ref{sec:escape} is a feature of the training loss: any account that locates a
transition in that curve has to rule this out first. Comparing training loss against
the closed-form floor does so in one line.

\paragraph{Bit-identical reruns.} Retraining a cell with the same seed under a
later version of the training script, which differs only in the density of probe
evaluations, reproduces the training loss exactly: 88 logged steps agree to all
recorded digits, including steps after the shortcut transition. The probe therefore
does not consume randomness shared with the training data stream, and the
cross-seed variation of \S\ref{sec:flip} contains no contribution from numerical
nondeterminism --- kernel selection, bfloat16 accumulation order, or dataloader
races. A run is replayable given its seed.

\paragraph{Two probe densities, one scale.} The probe evaluated during training
uses 200 documents and the terminal probe 400. Across the 50 runs of seeds 0 and 1 the two agree to 0.012 on average and 0.060 at worst in sign fraction, so trajectories
(\S\ref{sec:plateau}, App.~\ref{app:drift}) and terminal values
(Table~\ref{tab:stats}) are reported on the same scale. Both are computed over all
documents with edit domain and neither is gated on contrast-pair mass.

\paragraph{Why the label is deterministic.} An alternative way to make two rules
compete is to make the label itself ambiguous: return $v_{\mathrm{new}}$ with
probability $p$ and $v_{\mathrm{old}}$ with probability $1-p$. This does not
answer the question we are asking. The optimal policy becomes reproducing $p$, so
the objective pins the answer down and what becomes measurable is calibration
error rather than mechanism identity --- a model implementing either rule with the
right output distribution is optimal, and the two remain indistinguishable for the
same reason as before.

The aliased construction instead leaves mechanism choice underdetermined by the
loss while keeping the ground truth deterministic. Accuracy therefore remains a
meaningful convergence criterion: a model at 1.000 has solved the task, and the
question of which rule it uses is separate from the question of whether it has
converged. This separation is what makes \S\ref{sec:sat} and \S\ref{sec:flip}
independent claims rather than two readings of the same measurement.

\paragraph{The temporal invariant constrains the edit.} Update flags are
non-decreasing within a slot and same-valued occurrences are contiguous, so a
document in which the second of three $v_{\mathrm{old}}$ copies is replaced while
the first and third remain is outside the language. This is why the multiplicity
inversion of \S\ref{sec:probe} retains the earliest copy rather than an arbitrary
one: retaining a middle copy would interleave generations.

The choice has a side effect the readout depends on. Because
$v_{\mathrm{old}}$ occupies the first position of the queried slot in the base
document and still occupies it after the edit, any position-based account of the
readout contributes equally to both terms of Equation~\ref{eq:readout} and
cancels. This includes the specific failure mode of attention degrading to a
slot's first occurrence under off-distribution input, which the edit is otherwise
a natural candidate to trigger.

\subsection{Covariates}
\label{app:covar}

\emph{slots}: distinct (entity, attribute)
pairs per document. $R_{\mathrm{real}}$: surviving copies of $v_{\mathrm{old}}$
in the queried slot, below nominal because copies placed outside the document
canvas are discarded. \emph{domain}: fraction of documents with at least two
surviving copies, the upper bound on edit coverage before invariant re-checking.
\emph{support} and \emph{posCeil} are the realized $\Delta D$ interval and the
positional ceiling $1/\lvert\mathrm{supp}(\Delta D)\rvert$ of \S\ref{sec:ceiling}.
\emph{tail0-frac}: fraction with no update after the queried rebinding, matching
$(1-\rho)^{\Delta D}$ to within 1.5 points in every cell. This is a property of the
corpus, not the stratified accuracy on the same subset
(\S\ref{sec:sat}, App.~\ref{app:states}).

\begin{table}[h]
\centering
\caption{Covariates that shift with the axes, 1500 documents per cell. Columns are
defined in the text above.}
\label{tab:covar}
\begin{tabular}{rrrrrrrr}
\toprule
$R_{\mathrm{old}}$ & $\Delta D$ & support & posCeil & slots & $R_{\mathrm{real}}$ & domain & tail-0 \\
\midrule
3 & 2 & $[1,3]$ & 0.333 & 27.6 & 2.52 & 0.92 & 0.66 \\
3 & 3 & $[2,4]$ & 0.333 & 27.6 & 2.47 & 0.91 & 0.52 \\
3 & 5 & $[2,8]$ & 0.143 & 27.6 & 2.39 & 0.89 & 0.37 \\
3 & 8 & $[4,12]$ & 0.111 & 27.6 & 2.23 & 0.81 & 0.19 \\
3 & 16 & $[8,24]$ & 0.059 & 27.9 & 1.91 & 0.66 & 0.04 \\
5 & 2 & $[1,3]$ & 0.333 & 20.2 & 3.99 & 0.99 & 0.71 \\
5 & 3 & $[2,4]$ & 0.333 & 20.3 & 3.89 & 0.99 & 0.61 \\
5 & 5 & $[2,8]$ & 0.143 & 20.4 & 3.72 & 0.97 & 0.48 \\
5 & 8 & $[4,12]$ & 0.111 & 20.5 & 3.48 & 0.96 & 0.31 \\
5 & 16 & $[8,24]$ & 0.059 & 20.6 & 2.87 & 0.87 & 0.10 \\
8 & 2 & $[1,3]$ & 0.333 & 14.6 & 5.82 & 1.00 & 0.82 \\
8 & 3 & $[2,4]$ & 0.333 & 14.6 & 5.71 & 1.00 & 0.73 \\
8 & 5 & $[2,8]$ & 0.143 & 14.8 & 5.44 & 0.99 & 0.62 \\
8 & 8 & $[4,12]$ & 0.111 & 14.8 & 5.08 & 0.99 & 0.46 \\
8 & 16 & $[8,24]$ & 0.059 & 15.0 & 4.06 & 0.95 & 0.19 \\
12 & 2 & $[1,3]$ & 0.333 & 10.8 & 7.84 & 1.00 & 0.86 \\
12 & 3 & $[2,4]$ & 0.333 & 10.8 & 7.69 & 1.00 & 0.79 \\
12 & 5 & $[2,8]$ & 0.143 & 11.0 & 7.25 & 1.00 & 0.70 \\
12 & 8 & $[4,12]$ & 0.111 & 11.0 & 6.79 & 0.99 & 0.59 \\
12 & 16 & $[8,24]$ & 0.059 & 11.1 & 5.42 & 0.98 & 0.34 \\
16 & 2 & $[1,3]$ & 0.333 & 8.7 & 9.05 & 1.00 & 0.90 \\
16 & 3 & $[2,4]$ & 0.333 & 8.7 & 8.87 & 1.00 & 0.85 \\
16 & 5 & $[2,8]$ & 0.143 & 8.8 & 8.37 & 1.00 & 0.75 \\
16 & 8 & $[4,12]$ & 0.111 & 8.9 & 7.81 & 1.00 & 0.65 \\
16 & 16 & $[8,24]$ & 0.059 & 9.1 & 6.34 & 0.98 & 0.41 \\
\bottomrule
\end{tabular}

\end{table}

\subsection{Collision rates}
\label{app:cells}

Table~\ref{tab:collide} reports how often each candidate rule selects the ground
truth. A rule colliding with
the truth on every document cannot be separated from it by any observational
method --- no probe, no attention pattern, no output comparison --- because no
document exists on which the two disagree. This is what forces the causal edit of
\S\ref{sec:probe}.

\begin{table}[h]
\centering
\small
\caption{Fraction of documents on which each candidate rule selects the ground
truth, 1500 documents per cell. \textsc{rec} recency, \textsc{rar} rarity,
\textsc{frq} frequency, \textsc{pri} primacy, \textsc{glu} global last update,
\textsc{ltk} last value token before the query, \textsc{pos} best fixed offset
from the end.}
\label{tab:collide}
\begin{tabular}{rrrrrrrrr}
\toprule
$R_{\mathrm{old}}$ & $\Delta D$ & \textsc{rec} & \textsc{rar} & \textsc{frq} & \textsc{pri} & \textsc{glu} & \textsc{ltk} & \textsc{pos} \\
\midrule
3 & 2 & 1.000 & 1.000 & 0.085 & 0.000 & 0.651 & 0.000 & 0.341 \\
3 & 3 & 1.000 & 1.000 & 0.099 & 0.000 & 0.528 & 0.000 & 0.345 \\
3 & 5 & 1.000 & 1.000 & 0.120 & 0.000 & 0.359 & 0.000 & 0.163 \\
3 & 8 & 1.000 & 1.000 & 0.163 & 0.000 & 0.183 & 0.000 & 0.127 \\
3 & 16 & 1.000 & 1.000 & 0.309 & 0.000 & 0.042 & 0.000 & 0.070 \\
\midrule
5 & 2 & 1.000 & 1.000 & 0.024 & 0.000 & 0.722 & 0.000 & 0.339 \\
5 & 3 & 1.000 & 1.000 & 0.028 & 0.000 & 0.612 & 0.000 & 0.343 \\
5 & 5 & 1.000 & 1.000 & 0.026 & 0.000 & 0.487 & 0.000 & 0.163 \\
5 & 8 & 1.000 & 1.000 & 0.043 & 0.000 & 0.303 & 0.000 & 0.122 \\
5 & 16 & 1.000 & 1.000 & 0.130 & 0.000 & 0.101 & 0.000 & 0.071 \\
\midrule
8 & 2 & 1.000 & 1.000 & 0.005 & 0.000 & 0.831 & 0.000 & 0.353 \\
8 & 3 & 1.000 & 1.000 & 0.005 & 0.000 & 0.741 & 0.000 & 0.353 \\
8 & 5 & 1.000 & 1.000 & 0.009 & 0.000 & 0.626 & 0.000 & 0.155 \\
8 & 8 & 1.000 & 1.000 & 0.013 & 0.000 & 0.447 & 0.000 & 0.119 \\
8 & 16 & 1.000 & 1.000 & 0.049 & 0.000 & 0.195 & 0.000 & 0.069 \\
\midrule
12 & 2 & 1.000 & 1.000 & 0.001 & 0.000 & 0.884 & 0.000 & 0.351 \\
12 & 3 & 1.000 & 1.000 & 0.002 & 0.000 & 0.823 & 0.000 & 0.351 \\
12 & 5 & 1.000 & 1.000 & 0.006 & 0.000 & 0.720 & 0.000 & 0.153 \\
12 & 8 & 1.000 & 1.000 & 0.005 & 0.000 & 0.567 & 0.000 & 0.122 \\
12 & 16 & 1.000 & 1.000 & 0.021 & 0.000 & 0.333 & 0.000 & 0.070 \\
\midrule
16 & 2 & 1.000 & 1.000 & 0.002 & 0.000 & 0.900 & 0.000 & 0.337 \\
16 & 3 & 1.000 & 1.000 & 0.002 & 0.000 & 0.861 & 0.000 & 0.337 \\
16 & 5 & 1.000 & 1.000 & 0.002 & 0.000 & 0.755 & 0.000 & 0.161 \\
16 & 8 & 1.000 & 1.000 & 0.003 & 0.000 & 0.649 & 0.000 & 0.122 \\
16 & 16 & 1.000 & 1.000 & 0.025 & 0.000 & 0.442 & 0.000 & 0.074 \\
\bottomrule
\end{tabular}

\end{table}

\textsc{recency} and \textsc{rarity} both collide at 1.000 in every cell. This is
Equation~\ref{eq:alias} restated empirically: the identity holds document by
document, so the two are indistinguishable to any method that observes only inputs
and outputs.

\textsc{frequency} and \textsc{primacy} collide at 0.001--0.309 and 0.000
respectively. Both select $v_{\mathrm{old}}$ whenever more than one copy survives,
so they are the target's complement rather than its alias.
\textsc{frequency}'s rate matches $1 -$ domain from Table~\ref{tab:covar} cell by
cell (0.085 against 0.08 at $R_{\mathrm{old}}=3$, $\Delta D=2$; 0.002 against 0.00
at $R_{\mathrm{old}}=16$, $\Delta D=2$), since those are exactly the documents in
which one copy survives and both rules degenerate to \textsc{recency}.

\textsc{global last update} is the candidate whose rate varies most across the
grid, from 0.042 to 0.900, following $(1-\rho)^{\Delta D}$. It agrees with the
tail0-frac column of Table~\ref{tab:covar} to within 0.04 in every cell, the largest
gap being 0.033 at $R_{\mathrm{old}}=12$, $\Delta D=3$ --- two
independent computations of the same quantity, one from the closed form and one by
counting, on separate document samples.

\textsc{last value token} collides at 0.000 everywhere. Values are drawn without
replacement within a document, so the token immediately before the query is never
the answer. This removes a trivial copy baseline by construction rather than by
chance, and is one of the five invariants of App.~\ref{app:gen}.

\textsc{positional} collides at 0.069--0.353, tracking
$1/\lvert\mathrm{supp}(\Delta D)\rvert$ across the four support sizes:
0.337--0.353 against 0.333 at $\lvert\mathrm{supp}\rvert = 3$, 0.153--0.163
against 0.143 at 7, 0.119--0.127 against 0.111 at 9, and 0.069--0.074 against
0.059 at 17. The measured values sit one to two points above the analytic bound
because the empirical mode of a discrete uniform sample slightly exceeds
$1/\lvert\mathrm{supp}\rvert$. This is Equation~\ref{eq:posceil} arrived at by
counting rather than by derivation, and the agreement is a check on the derivation
that involves no trained model.

\section{Architecture choices}
\label{app:arch}

Three architectural choices could interact with the axes rather than sitting orthogonal to
them: the QK-normalization gain, the presence of the normalization itself, and depth. We
test the initial gain and depth against the main grid on subsets of cells. We do not
ablate the normalization itself; \S\ref{app:qk} states what that leaves open.

\subsection{The QK-normalization gain}
\label{app:qk}
Our attention applies RMS normalization to queries and keys per head before the dot
product. The normalization carries a learnable gain: one vector of length $d_h$ for
queries and one for keys in every layer, initialized to the constant $\gamma = 2.0$.
Writing $w^{(q)}, w^{(k)} \in \mathbb{R}^{d_h}$ for those vectors and $\hat q, \hat k$ for
the normalized query and key, each of unit root-mean-square per component so that
$\lVert \hat q \rVert = \lVert \hat k \rVert = \sqrt{d_h}$, the logit is
\begin{equation}
z = \frac{1}{\sqrt{d_h}} \sum_{i=1}^{d_h} w_i^{(q)} w_i^{(k)} \hat q_i \hat k_i,
\qquad
\lvert z \rvert \;\leq\; \Big( \max_i \lvert w_i^{(q)} w_i^{(k)} \rvert \Big) \sqrt{d_h}.
\label{eq:logitbound}
\end{equation}
At initialization every component equals $\gamma$, so this specializes to
$z = \gamma^2 \, \hat q \cdot \hat k / \sqrt{d_h}$ with
$\lvert z \rvert \leq \gamma^2 \sqrt{d_h}$, which at $d_h = 64$ is $8\gamma^2$. Two things
this bound is not. It is not a limit on what the architecture can express: the gains are
learnable and a model can raise the range by raising them, uniformly or on selected
dimensions, the latter requiring the query--key signal to be carried consistently on those
dimensions. And it is not what separates normalized from unnormalized attention, since at
fixed weights the unnormalized logit is also bounded --- by
$\lVert q \rVert \lVert k \rVert$, which is input-dependent rather than parametric. What
normalization does is remove the input's own norm from the logit scale and place that scale
under explicit parametric control.

The requirement comes from the sequence length. A document of $m$ statements is $4m+4$
tokens, 224 at $m = 55$. Suppose a head places fraction $p$ of its attention on one
position out of $n$ and the other $n-1$ logits are equal to $z_o$; then
\begin{equation}
z_t - z_o \;=\; \log \frac{p \, (n-1)}{1-p}.
\label{eq:gap}
\end{equation}
At $n = 224$ and $p = 0.9$ this is 7.6. The equal-distractor assumption is what makes this
a statement about a pairwise gap: in general the condition is
$z_t - \log\sum_{j \neq t} e^{z_j} \geq \log\frac{p}{1-p}$, and a uniform gap of 7.6 is
sufficient rather than necessary, since a head that has learned to suppress particular
positions can reach $p = 0.9$ with a smaller maximum gap. We use the equal-distractor case
because it describes a head that has not yet learned which positions to suppress, which is
the situation during formation.

At initialization the gap decomposes as
$\Delta z = 8\gamma^2 \, (\cos\theta_t - \cos\theta_o)$, writing $\cos\theta$ for
$\hat q \cdot \hat k / d_h \in [-1, 1]$. Reaching 7.6 therefore requires a
target--distractor cosine gap of 0.95 at $\gamma = 1$ against 0.24 at $\gamma = 2$. The
first is attainable only near perfect alignment on the target together with near
orthogonality on the other 223 positions, and it must be found while the same head also
needs intermediate confidences on other inputs; the second leaves most of the range for
graded preferences.

The interaction with our task is specific, and it is an expectation the arm tests rather
than a fact we assert. Both mechanisms need selective routing. The difference is that the
positional shortcut can reuse one input-independent offset pattern on every document, which
RoPE supplies directly, while retrieval needs a content match between the query and the
correct slot and therefore an input-dependent alignment. We expect a restricted initial
logit scale to impede the second more than the first, which would make it a confound on the
axis we measure; that is what the arm below is for. We set $\gamma = 2$ so that
Equation~\ref{eq:gap} is satisfied with margin at every $n \leq 224$.

\paragraph{The gain is checked at one point, in two cells.}
We ran $\gamma = 1$ at $R_{\text{old}} = 3, \Delta D = 5$ and $R_{\text{old}} = 16,
\Delta D = 2$, two seeds each, with every other field of the corpus, model and training
configuration verified identical to the main grid. The two cells differ fivefold in escape
time at $\gamma = 2$ (gate at 5000 and 1000 steps), so the arm compares a slowly forming
circuit against a fast one.

The constraint binds only on the slow cell. At $R_{\text{old}} = 16$ the gate is crossed at
step 1000 under both gains and both seeds, with accuracy and the copy diagnostic both at
1.000 from the 1000-step evaluation onward. At $R_{\text{old}} = 3$ neither seed crosses it
within 16000 steps: the copy diagnostic terminates at 0.132 and 0.244 against 1.000 at
$\gamma = 2$, and accuracy at 0.092 and 0.194 --- 65\% and 136\% of the positional ceiling
of 0.143.

\paragraph{The logit bound does not explain that asymmetry.} The bound of
Equation~\ref{eq:logitbound} is set by the learned gains, so we measured them at the final
checkpoint of all four runs and of the main grid (Table~\ref{tab:gain}). Three findings, and
the first two are against the account above. At $\gamma = 1$ the terminal bound is 20.5 to
36.9, three to five times the 7.6 that Equation~\ref{eq:gap} requires, and the
$R_{\mathrm{old}}=3$ seed-1 run reaches 36.9 while never forming the circuit --- higher than
several main-grid runs that form it by step 1000. The gains do not stay at their
initialization: the mean falls while the largest component roughly doubles, so the
distribution spreads and the bound rises even as the average gain drops. And
$R_{\mathrm{old}}=16$ at $\gamma = 1$ crosses the gate at step 1000, when the gains cannot
yet have moved far, so a bound near the initial 8.0 is sufficient for the circuit in that
cell. Whatever prevents formation at $R_{\mathrm{old}}=3$, $\gamma = 1$, it is not that the
attainable logit range is too small to express confident retrieval.

We therefore report the arm as an observation without a mechanism. What it establishes is
that the initial gain is not a free parameter with respect to which cells can be analyzed;
what it does not establish is why. Distinguishing the remaining accounts --- that the
constraint binds early and the cosine schedule has decayed by the time the gains spread,
that the initial region satisfying Equation~\ref{eq:gap} is small enough to be hard to find
from a cold start, or that the gain interacts with formation in a way the logit range does
not capture --- requires the gain trajectory rather than its endpoint, which our
checkpoints of this arm do not carry.

\begin{table}[h]
\centering
\small
\caption{Terminal QK-norm gains. \emph{mean}, \emph{max} and \emph{std} are over all
$2 \, d_h \, n_{\mathrm{layer}} = 1024$ components; \emph{bound} is
$\max_i \lvert w^{(q)}_i w^{(k)}_i \rvert \sqrt{d_h}$, the right-hand side of
Equation~\ref{eq:logitbound}, against the 7.6 that Equation~\ref{eq:gap} requires at
$n = 224$, $p = 0.9$. Initial bounds are 8.0 at $\gamma = 1$ and 32.0 at $\gamma = 2$.
\emph{escape} is the 1000-step gate. Main-grid rows are the range over the 50 runs of seeds
0 and 1.}
\label{tab:gain}
\begin{tabular}{llrrrrr}
\toprule
run & $\gamma$ & escape & mean & max & std & bound \\
\midrule
$R_{\mathrm{old}}=3, \Delta D=5$, s0   & 1.0 & none & 0.848 & 1.703 & 0.345 & 22.9 \\
$R_{\mathrm{old}}=3, \Delta D=5$, s1   & 1.0 & none & 0.958 & 2.156 & 0.425 & 36.9 \\
$R_{\mathrm{old}}=16, \Delta D=2$, s0  & 1.0 & 1000 & 1.067 & 1.602 & 0.248 & 20.5 \\
$R_{\mathrm{old}}=16, \Delta D=2$, s1  & 1.0 & 1000 & 1.023 & 1.727 & 0.274 & 23.6 \\
\addlinespace
main grid, 50 runs                     & 2.0 & 1000--10000 & 1.47--1.74 & 2.11--2.55 & 0.15--0.35 & 33--52 \\
\bottomrule
\end{tabular}
\end{table}

\paragraph{The initialization is not binding at $\gamma = 2$.} Across the 50 main-grid runs
the terminal mean gain is 1.47 to 1.74, below its initialization of 2.0, and it falls
monotonically with budget in the two cells where we varied that: 1.832, 1.621 and 1.475 at
4000, 16000 and 32000 steps at $R_{\mathrm{old}}=16$, $\Delta D=2$, and 1.660 against 1.451
at 16000 and 32000 at $R_{\mathrm{old}}=3$, $\Delta D=2$. Optimization moves the gain down
from 2.0 rather than up, so at $\gamma = 2$ the initialization sits above what the task
uses and does not constrain the runs the paper analyzes. This is the sense in which the
choice is conservative.

\begin{table}[ht]
\centering
\caption{The $\gamma = 1$ arm. \emph{escape} is the first evaluation at which the copy
diagnostic exceeds 0.95; \emph{none} means it is never crossed within 16000 steps.
\emph{mass} is mean contrast-pair mass and \emph{median} is ungated, both as in
Table~\ref{tab:stats}. Readouts are omitted at $R_{\mathrm{old}}=3$, $\gamma=1$:
mass is 0.15 there, with 0.11 and 0.02 of documents clearing the per-document gate, so
those two runs report on circuit formation and not on rule attribution.}
\label{tab:gamma}
\begin{tabular}{llrrrrrr}
\toprule
cell & $\gamma$ & escape & acc & copy & mass & median & frac+ \\
\midrule
$R_{\mathrm{old}}=3, \Delta D=5$, s0 & 2.0 & 5000 & 1.000 & 1.000 & 1.00 & $-0.087$ & 0.126 \\
$R_{\mathrm{old}}=3, \Delta D=5$, s0 & 1.0 & none & 0.092 & 0.132 & 0.15 & --- & --- \\
$R_{\mathrm{old}}=3, \Delta D=5$, s1 & 2.0 & 5000 & 1.000 & 1.000 & 0.66 & $+1.964$ & 0.972 \\
$R_{\mathrm{old}}=3, \Delta D=5$, s1 & 1.0 & none & 0.194 & 0.244 & 0.15 & --- & --- \\
\addlinespace
$R_{\mathrm{old}}=16, \Delta D=2$, s0 & 2.0 & 1000 & 1.000 & 1.000 & 0.98 & $+1.827$ & 0.998 \\
$R_{\mathrm{old}}=16, \Delta D=2$, s0 & 1.0 & 1000 & 1.000 & 1.000 & 0.97 & $+2.095$ & 0.993 \\
$R_{\mathrm{old}}=16, \Delta D=2$, s1 & 2.0 & 1000 & 1.000 & 1.000 & 1.00 & $+0.113$ & 0.736 \\
$R_{\mathrm{old}}=16, \Delta D=2$, s1 & 1.0 & 1000 & 1.000 & 1.000 & 0.89 & $+2.143$ & 0.987 \\
\bottomrule
\end{tabular}
\end{table}

\paragraph{Two consequences for the grid.} First, the composition of the analyzed grid
depends on $\gamma$: at $\gamma = 1$ the tested $R_{\mathrm{old}}=3, \Delta D=5$ cell leaves
it the way the $R_{\mathrm{old}}=2$ column does at $\gamma = 2$
(App.~\ref{app:posneg}), so that column would no longer be fully analyzable. Second, the
readout at $R_{\mathrm{old}}=3$ is not reportable at $\gamma = 1$ in either seed:
contrast-pair mass is 0.15 with 0.11 and 0.02 of documents clearing the per-document gate,
so those two runs enter this appendix as a statement about circuit formation and not about
rule attribution. At $R_{\mathrm{old}}=16$ (Table~\ref{tab:gamma}) the gated medians at
$\gamma = 1$ are $+2.05$ and $+1.84$ with sign fractions 0.99 and 0.99; the cross-seed
spread of that cell at $\gamma = 2$ is of the same order as any difference between the
gains, so we do not read a readout effect off two seeds.

\paragraph{We did not remove the normalization.} The arm above varies the initial gain at
fixed architecture; it does not separate the effect of normalizing from the effect of the
initial scale. The third condition --- normalization removed entirely, which is not
$\gamma = 0$ since the gain is a learnable vector and this removes it along with
$2 d_h n_{\mathrm{layer}} = 1024$ parameters --- is not run. Under the initialization of
\S\ref{sec:inv}, and by the calculation above rather than by measurement, the unnormalized
attention logit has standard deviation near 1 at the start, close to what $\gamma = 1$
gives, so the two conditions may not be far apart at initialization. What distinguishes
them is the parameterization: with normalization the input-wise logit range is set by the
gain vectors and is independent of the raw query and key norms, while without it those norms
provide an additional input-dependent route.

\subsection{Depth}
\label{app:depth}

We repeated the $\Delta D = 5$ column at 4 and 12 layers, $R_{\mathrm{old}} \in
\{3, 8, 16\}$, two seeds each; the 8-layer entries are the main grid's. Width is fixed at
$d_{\mathrm{model}} = 512$, so depth and capacity covary (13.4M, 26.1M and 38.7M
parameters). The corpus configuration is identical to the main grid, so at a given seed
the documents are the same to the byte and the comparison is paired at the document
level, which the arm of App.~\ref{app:fixband} cannot be.

\paragraph{Both positive claims survive.} All twelve runs reach in-distribution accuracy
1.000, the copy diagnostic 1.000, and the \textsc{retrieval} state, so depth does not
change which cells are analyzable --- unlike the gain of App.~\ref{app:qk}. On the
1000-step gate, escape is 4000--6000 steps at $R_{\mathrm{old}}=3$, 1000--2000 at $8$ and
1000 at $16$ across both new depths, so the ordering of \S\ref{sec:escape} holds at 4, 8
and 12 layers, with $8$ and $16$ on the gate's floor as they are in the grid. The
within-cell gradient of \S\ref{sec:covar} runs in the direction of each cell's own effect
in all twelve runs, including the two whose cell median is negative. Pooled over seeds, $R_{\mathrm{old}}=3$ has the lowest sign fraction at every depth (0.59,
0.55 and 0.42 at 4, 8 and 12 layers), while the ordering between $8$ and $16$ is not
consistent across depths (0.99 against 0.97 at four layers, 0.97 against 0.79 at eight,
0.81 against 0.99 at twelve).

\paragraph{The loss derivative does not sharpen the separation here.} In the grid the
derivative peak resolves four levels of $R_{\mathrm{old}}$ where the gate resolves three
(App.~\ref{app:states}). It does not do so in this arm. Table~\ref{tab:depthpeak} gives the
peaks. Four of the twelve runs are unusable because the schedule's final decay exceeds the
escape peak, and those four remove $R_{\mathrm{old}}=16$ entirely at twelve layers. What
the usable runs support is the coarse ordering with a wide margin ---
$R_{\mathrm{old}}=3$ at 3400--5100 steps against 400--1200 elsewhere --- and not the finer
separation: at four layers the usable $8$ and $16$ peaks are both 400, and at twelve layers
there is no usable $16$ peak to compare against $8$. The gate is coarser but complete, and
gives the same coarse ordering at both depths. Each row here is one cell per seed against
five in the grid, so the loss of resolution is consistent with reduced power rather than
with an effect of depth, and the arm cannot separate the two.

\begin{table}[h]
\centering
\small
\caption{Loss-derivative peak position in the depth arm, both seeds, on the 100-step
recording grid. $^{\dagger}$ marks runs where the cosine schedule's final decay exceeds the
escape peak, so the search returns a position inside the decay rather than at formation
(\S\ref{sec:escape}); those values are not usable. The 8-layer rows are the main grid's
(Table~\ref{tab:escape}).}
\label{tab:depthpeak}
\begin{tabular}{rr rr}
\toprule
layers & $R_{\mathrm{old}}$ & seed 0 & seed 1 \\
\midrule
4  & 3  & 3400 & 5100 \\
4  & 8  & 400  & 500$^{\dagger}$ \\
4  & 16 & 300$^{\dagger}$ & 400 \\
\addlinespace
8  & 3  & 4200 & 4600 \\
8  & 8  & 800  & 900 \\
8  & 16 & 400  & 400 \\
\addlinespace
12 & 3  & 5100 & 4400 \\
12 & 8  & 1200 & 1000 \\
12 & 16 & 400$^{\dagger}$ & 400$^{\dagger}$ \\
\bottomrule
\end{tabular}
\end{table}

\paragraph{The per-cell readout moves with depth, without an ordering in it.} At
$R_{\mathrm{old}}=3$ the sign fraction is 0.91, 0.13 and 0.12 in seed 0 at 4, 8 and 12
layers and 0.26, 0.97 and 0.72 in seed 1. Seed 0 falls with depth and seed 1 rises then
falls, so the two share no trend and place opposite sides of 0.5 at the same depths. The
cell median at $R_{\mathrm{old}}=3$, seed 0 moves from $+2.858$ at 4 layers to $-0.058$ at
12 --- a change of sign, not merely of magnitude. The largest movement with depth at fixed
seed is 0.79 in sign fraction; the largest with seed at fixed depth is 0.65 in this arm and
0.846 in the same cell of the grid. With two seeds per depth the arm cannot rank the two
terms, so we claim only that the movement associated with changing depth and capacity
together is of the same order as the movement associated with reseeding --- which is what
\S\ref{sec:flat} predicts for a direction the objective does not constrain.

\paragraph{What this arm does not settle.} Depth and capacity are not separated, and
because the readout has no ordering in depth we cannot say whether a deeper model would
eventually stabilize it. The arm is one column, chosen when two seeds were available, at which point
$\Delta D = 5$ carried the widest cross-seed spread at $R_{\mathrm{old}}=3$ (0.846);
with the third seed it is second to $\Delta D = 8$ (0.845 against 0.879,
Table~\ref{tab:spread}). Either way the readout comparison at that row is among the
least conclusive available, and the escape comparison is unaffected.

\section{Edit validity in detail}
\label{app:oracle}

\paragraph{The five edit invariants.} Token count, answer position, ground-truth
answer and statement count are preserved by construction. The fifth --- the
distance from the final queried statement to its nearest same-slot antecedent ---
is not automatic: an edit changing it would covary with $\Delta D$ in a way absent
from training. We verify all five document by document and discard violations,
which is one of two reasons edit coverage is below 1.00 (the other is that the
inversion requires at least two surviving copies of $v_{\mathrm{old}}$).

\paragraph{Idealized predictors.} A predictor assigning logit $\kappa$ to the
target rule's prediction returns $\Delta = 2\kappa$; one assigning it to the
ground truth returns $\Delta = 0$ identically. At $\kappa = 8$ we measure $+16.0$
and $|\Delta| < 10^{-9}$ in every cell. This is a check on the construction
rather than a result: the values follow from Equation~\ref{eq:readout}, and
measuring them confirms the edit has no side effect on the contrast pair.

\paragraph{A trained model with no mechanism.} Before the circuit forms, $\Delta$
is $+0.000$ with 0.49 of documents positive at step 2000 and $-0.002$ with 0.40 at
step 4000 ($R_{\mathrm{old}} = 3$, $\Delta D = 2$). This is stronger than the
idealized check, because the network is trained rather than constructed --- if the
edit perturbed logits on its own, a model with nothing to move would not return
zero.

\paragraph{The filler-slot control.} The same inversion applied to a non-queried
slot perturbs the same number of tokens while leaving the queried slot untouched,
and reads $-0.009$ to $+0.015$ across the grid --- below 1\% of the targeted
readout wherever that readout exceeds 1 nat. In the five $R_{\mathrm{old}}=3$ cells of
seed 0, where the median is 0.011--0.087 nats, the control is opposite in sign in all
five and at most 5\% of the median, so it is not producing a spurious effect in the
same direction there. Two cells elsewhere in the grid do not admit that argument: at
$R_{\mathrm{old}}=8$, $\Delta D=2$ in seed 1 the control is $-0.009$ against a gated
median of $-0.006$, and at $8, 16$ in seed 2 it is $-0.005$ against $-0.012$, the signs
agreeing in both. Both are among the four cells whose gated median is negative at
$R_{\mathrm{old}} \geq 5$ (\S\ref{sec:sign}), which is a reason to read those four as
edit noise rather than as accumulation. The most extreme case is outside the grid:
$-0.020$ against a median of $-0.055$ with the signs agreeing, at
$R_{\mathrm{old}}=8$, $\Delta D=2$ in the
$p_{\mathrm{update}}$ arm, whose readout we therefore report as uninterpretable
(App.~\ref{app:pupd}). Per-cell control values: Table~\ref{tab:stats}.

\paragraph{The off-distribution mass gate.} The inversion makes the answer value
occur more than once before the query, which never happens during training. We
therefore report the probability mass on $\{v^\star, v_{\mathrm{truth}}\}$ and
treat readouts below 0.5 as invalid, applying the gate per document so that a cell
mean cannot conceal documents on which the model has left the task. Measured mass is 0.59--1.00 across the analyzed grid, falling as the effect grows
(\S\ref{sec:degrade}); in one cell the gate moves the median by
1\% at 4000 steps and 16\% at 32000, and the mechanism is a growing subpopulation
of near-saturated documents rather than a longer tail (App.~\ref{app:drift}).

\paragraph{Why coverage falls with $\Delta D$.} The edit requires at least two
copies of $v_{\mathrm{old}}$ to survive into the document, giving a domain of
0.69--1.00 after all five invariants are re-checked, or 0.66--1.00 counting only
surviving copies (Table~\ref{tab:covar}). Coverage falls with $\Delta D$ because
copies placed outside the document canvas are discarded, so cells at low
$R_{\mathrm{old}}$ and high $\Delta D$ condition on a subset with higher realized
redundancy. \S\ref{sec:covar} tests this with a within-cell split.

\section{The positional shortcut}
\label{app:pos}
The shortcut's ceiling follows in closed form from the token layout; the cells that
occupy it longest are the excluded $R_{\mathrm{old}}=2$ column, so the derivation and
those cells are reported together.

\subsection{Derivation of the ceiling}
\label{app:posceil}
Each statement occupies exactly four tokens, so statement $j$ (zero-indexed)
spans token positions $4j$ through $4j+3$ with its value token at $4j+2$. The
query appends four tokens, giving a document of $4m+4$ tokens for $m$ statements.

If the query follows $\Delta D$ statements after the rebinding of the queried
slot, that rebinding is statement $m - \Delta D - 1$ and its value token sits at
position $4(m - \Delta D - 1) + 2 = 4m - 4\Delta D - 2$. Counting from the end of
the sequence,
\begin{equation}
\text{offset of the answer value from the last token}
= (4m + 4) - (4m - 4\Delta D - 2) = 4\Delta D + 6.
\label{eq:offset}
\end{equation}

Equation~\ref{eq:offset} does not contain $m$. The answer's position measured
backward from the end depends only on $\Delta D$; document length cancels.

Consider a rule that ignores the query tokens and emits the value token at a
fixed offset $k$ from the end. By Equation~\ref{eq:offset} it is correct exactly
when $4\Delta D + 6 = k$, that is when $\Delta D$ takes one particular value.
Maximizing over $k$,
\begin{equation}
\mathrm{posCeil} = \max_d \Pr[\Delta D = d]
= \frac{1}{\lvert \mathrm{supp}(\Delta D) \rvert},
\label{eq:ceil}
\end{equation}
the second equality holding because $\Delta D$ is uniform on its support.

\subsection{What the bound requires}

Equation~\ref{eq:ceil} is an upper bound over positional rules only if $\Delta D$
cannot be predicted from anything the model observes. A rule conditioning on an
observable feature correlated with $\Delta D$ could exceed it.

Two features are candidates. Document length is uninformative by
Equation~\ref{eq:offset}: the offset is length-independent, so length carries no
signal about where the answer lies. Empirically the correlation between realized
$\Delta D$ and document length is at most 0.043 in every cell
(Table~\ref{tab:inv}). The support is also uniform in practice, not merely by design: at
$\lvert\mathrm{supp}(\Delta D)\rvert = 17$ the seventeen values carry 0.050--0.063
of the probability each over 3000 documents, a max-to-min ratio of 1.26 where
sampling 3000 documents into 17 bins gives a standard error of 0.004 per bin. (The
mean over the seventeen is $1/17$ identically and is not a check.) This makes the
second equality in Equation~\ref{eq:ceil} exact
rather than approximate, and it also means no positional offset is preferred over
another at wide support --- a fact we return to in App.~\ref{app:posneg}. Local token content is uninformative because the queried
slot's statements are drawn from the same vocabularies and emitted in the same
format as filler statements, and the unmarked language contains no update token
to distinguish a rebinding from an initial binding.

One constraint deserves note. Length must satisfy
$n_{\mathrm{stmts}} \geq R_{\mathrm{old}} + \Delta D_{\max} + 5$, which at the
widest cell ($R_{\mathrm{old}}=16$, $\Delta D_{\max}=24$) requires
$n_{\mathrm{stmts}} \geq 45$ --- exactly the lower end of $U[45,55]$. The
constraint is satisfied without truncating the $\Delta D$ distribution, but it
binds at the boundary: widening either axis further would induce a correlation
between $\Delta D$ and length and thereby break the bound rather than merely
loosen it, since a rule could then read length to predict $\Delta D$.

The queried slot is not, however, perfectly unmarked: its copies are more
dispersed than a filler slot's by 2--22\% (App.~\ref{app:gen}). A rule that
located the slot from this dispersion and then read its final value would exceed
Equation~\ref{eq:ceil}, but it would also require the slot-matching
computation that the copy diagnostic detects, so it is not a positional rule in the sense bounded here.

\subsection{Observed plateau heights}

\begin{table}[ht]
\centering
\caption{Positional ceiling against observed plateau accuracy. Accuracy is taken
at the last evaluation before the copy diagnostic leaves chance, except in the last
row ($R_{\mathrm{old}}=2$, $\Delta D=5$), where it never leaves chance and the value
is the terminal one at 12000 steps.}
\label{tab:ceiling}
\begin{tabular}{rrrrrrrl}
\toprule
$R_{\mathrm{old}}$ & $\Delta D$ & support & $\lvert\mathrm{supp}\rvert$
& posCeil & acc & copy & ratio \\
\midrule
2 & 2  & $[1,3]$  & 3 & 0.333 & 0.3255 & 0.002 & 98\% \\
2 & 3  & $[2,4]$  & 3 & 0.333 & 0.3253 & 0.001 & 98\% \\
3 & 5  & $[2,8]$  & 7 & 0.143 & 0.1045 & 0.002 & 73\% \\
2 & 8  & $[4,12]$ & 9 & 0.111 & 0.0773 & 0.046 & 70\% \\
2 & 5  & $[2,8]$  & 7 & 0.143 & 0.0018 & 0.003 & 1\% \\
\bottomrule
\end{tabular}
\end{table}

Four of these five cells are at $R_{\mathrm{old}}=2$ and therefore outside the
analyzed grid (\S\ref{sec:axes}); they are the cells that occupy the plateau longest
and so the only ones where its height is measurable for thousands of steps.

The two cells at $\lvert\mathrm{supp}\rvert = 3$ saturate the bound and agree with
each other to 0.0002 despite drawing from different bands, an internal replicate of
the closed-form account: same ceiling, independent cells. The three wider-support cells attain 70--73\% of
their ceilings or less, so occupancy falls with support width while the bound
itself falls. We report the ratio rather than the raw accuracy for this reason.

The row is non-monotone in band width: 98\%, 98\%, 1\%, 70\% at
$\Delta D = 2, 3, 5, 8$ with $R_{\mathrm{old}}=2$. The $\Delta D = 5$ cell reaches
neither the positional ceiling nor retrieval within 12000 steps, terminating at
0.0018 accuracy with a copy diagnostic of 0.003. Its neighbour at $\Delta D = 8$
occupies 70\% of a lower ceiling. We have no account of why one cell fails to find
either solution while both its neighbours find one, and we report it because it
bears on \S\ref{sec:escape}: whatever sets escape time, it is not a monotone
function of shortcut payoff alone.

\subsection{The \texorpdfstring{$R_{\mathrm{old}} = 2$}{Rold = 2} cells}
\label{app:posneg}

These five cells are excluded from the main grid because four of them never acquire the retrieval circuit within our budget; three of those four terminate in the \textsc{positional} state and one in \textsc{neither}. We report them because they are the clearest
evidence for the two-phase account of \S\ref{sec:dyn}: models that terminate on
the positional shortcut, and a probe reading taken on those models that would be
misread as a rule attribution.

\paragraph{Terminal states.} All five were trained for 12000 steps under the
same recipe as the main grid.

\begin{table}[h]
\centering
\caption{The excluded $R_{\mathrm{old}}=2$ column at 12000 steps, seed 0. Four of the
five never acquire the retrieval circuit within the budget: three terminate
\textsc{positional} and one \textsc{neither}. The probe reading on the three
positional runs is the artifact analyzed in \S\ref{sec:plateau}.}
\label{tab:posneg}
\begin{tabular}{rrrrrrl}
\toprule
$\Delta D$ & support & posCeil & acc & copy & $\Delta$ / sign frac & state \\
\midrule
2  & $[1,3]$  & 0.333 & 0.3255 & 0.002 & $-0.47$ / 0.05 & positional, 98\% of ceiling \\
3  & $[2,4]$  & 0.333 & 0.3253 & 0.001 & $-0.44$ / 0.01 & positional, 98\% \\
5  & $[2,8]$  & 0.143 & 0.0018 & 0.003 & $-0.00$ / 0.22 & neither, 1\% \\
8  & $[4,12]$ & 0.111 & 0.0773 & 0.046 & $-1.30$ / 0.02 & positional, 70\% \\
16 & $[8,24]$ & 0.059 & 1.0000 & 0.999 & $+0.33$ / 0.72 & retrieval, escaped at 5000 \\
\bottomrule
\end{tabular}
\end{table}

Two features of this table matter beyond the exclusion decision. The row is
non-monotone in band width, analyzed in \S\ref{app:posceil}. And the three
positional rows return $\Delta$ with the same sign and much higher internal
consistency than any low-$R_{\mathrm{old}}$ cell in the main grid --- 0.01 to
0.05 of documents positive, against 0.09 to 0.41 at $R_{\mathrm{old}}=3$.

\paragraph{Why the reading is not a rule attribution.} Taken at face value,
$-0.47$, $-1.30$ and a third pilot measurement of $-0.94$ (sign fraction 0.03,
at $R_{\mathrm{old}}=3$, $\Delta D=5$ under an earlier vocabulary setting of 128
values, so not directly comparable) describe a strongly frequency-type region at
low redundancy: multiplicity boosting the value the model should suppress,
precisely the majority-rule pattern the knowledge-conflict literature reports.
The models producing these numbers have copy diagnostics of 0.001 to 0.046. They
have no retrieval mechanism to attribute. Whatever the edit is doing to their
logits, it is not selecting between a recency rule and a rarity rule, because
neither is implemented.

\paragraph{A candidate account, and why we cannot test it here.} The natural
account is positional. The inversion rewrites $R-1$ copies of $v_{\mathrm{old}}$
as copies of $v_{\mathrm{new}}$ at their original statement positions; a
positional rule reads the value token at a fixed offset $4\Delta D + 6$ from the
end (App.~\ref{app:posceil}), so when that offset lands on a rewritten copy the
rule's output flips toward $v_{\mathrm{new}}$, and since
$v^\star = v_{\mathrm{old}}$ for this edit, $\Delta$ is negative.

Testing this requires knowing which offset the model settled on, and that is
identified only where the $\Delta D$ distribution has a distinct mode. It does
not: $\Delta D$ is uniform on its support by construction, so at
$\lvert\mathrm{supp}\rvert = 17$ each of the seventeen values carries
0.050--0.063 of the probability and the empirical mode moves between samples.
Computing the overlap between the rewritten positions and the modal offset gives
0.345, 0.339, 0.161, 0.120 and 0.091 at $\Delta D = 2, 3, 5, 8, 16$, but only the
first two rest on an identified offset ($\lvert\mathrm{supp}\rvert = 3$, where the
mode carries a third of the mass). Those two are consistent with the account ---
the highest overlaps, with $\lvert\Delta\rvert$ of 0.47 and 0.44 --- and two
points establish nothing.

We therefore report the plateau readings as an artifact whose mechanism we have not
identified, and note that identifying it would require reading the offset off the
model rather than inferring it from the data distribution. The argument of
\S\ref{sec:plateau} does not depend on that identification: it needs only the copy
diagnostics.

\paragraph{What we do with these cells.} Nothing in the main results depends on
them; they are excluded before any readout is reported. The reason for excluding
by circuit formation rather than by accuracy is that accuracy does not separate
them from the analyzed grid in the way one might expect. The
$\Delta D = 16$ cell reaches 1.000 and enters no analysis only because
$R_{\mathrm{old}}=2$ is outside the grid --- had it been at $R_{\mathrm{old}}=3$
it would have been analyzed, and correctly so. Conversely a cell at 0.3255
accuracy is not a failed run in any ordinary sense: it has converged, to a
different function. This is why the gate in \S\ref{sec:dyn} is on the copy
diagnostic and not on loss or accuracy.

\section{Per-cell state classification}
\label{app:states}

Every readout in this paper is gated on circuit formation, so we report the gate
per cell. A run is classified RETRIEVAL if the copy diagnostic exceeds 0.95 and
in-distribution accuracy exceeds 0.99 at the final checkpoint, POSITIONAL if the
copy diagnostic stays below 0.5 and accuracy is at least half the positional
ceiling (Eq.~\ref{eq:ceil}), with the attained fraction of that ceiling reported
alongside the label, and NEITHER otherwise. The three states are disjoint on the runs
that enter the main results: across the analyzed grid and the $R_{\mathrm{old}}=2$
column, the copy diagnostic at each run's final checkpoint is either at most 0.046 or
at least 0.999, so any threshold in $[0.05, 0.95]$ returns the same labels. The
$\gamma = 1$ runs of App.~\ref{app:qk} sit between those bounds (0.132 and 0.244);
they are outside the analyzed grid, and the label we report for them is the attained fraction of the positional ceiling rather than a threshold-insensitive state.

All 75 runs across the three seeds terminate in the RETRIEVAL state.
Accuracy is 1.000 to four decimals in every cell except $R_{\text{old}} = 3,
\Delta D = 16$ of seed 0 (0.9998) and $R_{\text{old}} = 8, \Delta D = 16$ of seed 2 (0.99975), which is why the main text states the bound as $\geq 0.999$; accuracy restricted to documents with no update after the queried
rebinding is 1.000 in every cell, matching unstratified accuracy to four decimals;
and the copy diagnostic is 1.000 in every cell. Escape steps on the 1000-step gate
are the \texttt{gate} column of Table~\ref{tab:escape}, which reports both seeds
and both localizations.

\paragraph{The gate is not doing selection work within the analyzed grid.} All 75
runs pass, so the classification excludes nothing here. Its function is to
exclude the $R_{\mathrm{old}}=2$ column (App.~\ref{app:posneg}, where three of
five cells terminate \textsc{positional}) and to timestamp the readouts: because
escape is as late as step 10000 at $R_{\mathrm{old}}=3$, a probe applied at step
4000 would have returned a rule attribution for a model with no rule
(\S\ref{sec:plateau}).

\begin{table}[ht]
\centering
\small
\caption{Escape step located two ways, both seeds. \emph{gate} is the first
evaluation at which the copy diagnostic exceeds 0.95 and is resolved only to the
1000-step evaluation grid. \emph{peak} is the position of the single interior
maximum of $d\,\mathrm{loss}/d\log\mathrm{step}$, recorded every 100 steps and
involving no threshold, with its height $h$. The peak resolves escape ten times
more finely and separates four levels of $R_{\mathrm{old}}$ ($3$, $5$, $8$ and
$\{12,16\}$) where the gate separates three. Peak height increases with escape step
(Spearman $\rho = 0.933$ over these 50 runs); the cosine schedule's final decay exceeds
the escape peak in the three cells marked $^{\dagger}$, so the
unthresholded argument is weak there and we rely on the gate. Nineteen of the 50 runs share a peak position of 400 steps, the floor of the
100-step recording grid; the Spearman coefficient of \S\ref{sec:escape} uses
average ranks over that tie group. The table covers seeds 0 and 1; seed 2's gate and peak
columns are in the text below. No seed-1 run is affected by the cosine tail.}
\label{tab:escape}
\begin{tabular}{rr rrr rrr}
\toprule
& & \multicolumn{3}{c}{seed 0} & \multicolumn{3}{c}{seed 1} \\
\cmidrule(lr){3-5} \cmidrule(lr){6-8}
$R_{\mathrm{old}}$ & $\Delta D$ & gate & peak & $h$ & gate & peak & $h$ \\
\midrule
3 & 2 & 8000 & 7500 & 9.86 & 7000 & 6100 & 8.12 \\
3 & 3 & 6000 & 5600 & 6.66 & 5000 & 4600 & 6.98 \\
3 & 5 & 5000 & 4200 & 7.15 & 5000 & 4600 & 5.77 \\
3 & 8 & 7000 & 6800 & 8.80 & 6000 & 5600 & 7.55 \\
3 & 16 & 4000 & 3400 & 5.11 & 10000 & 9600 & 14.48 \\
\addlinespace
5 & 2 & 2000 & 1000 & 3.36 & 3000 & 2800 & 6.80 \\
5 & 3 & 1000 & 1000 & 3.10 & 3000 & 2300 & 5.59 \\
5 & 5 & 1000 & 800 & 2.71 & 3000 & 2000 & 4.90 \\
5 & 8 & 4000 & 2900 & 6.40 & 4000 & 2900 & 7.99 \\
5 & 16 & 4000 & 3000 & 9.37 & 4000 & 3300 & 9.30 \\
\addlinespace
8 & 2 & 1000 & 800 & 3.46 & 1000 & 900 & 4.13 \\
8 & 3 & 1000 & 600 & 2.80 & 2000 & 1100 & 4.77 \\
8 & 5 & 1000 & 800 & 3.42 & 1000 & 900 & 3.97 \\
8 & 8 & 1000 & 700 & 3.19 & 2000 & 1600 & 6.36 \\
8 & 16 & 1000 & 800 & 3.48 & 2000 & 1400 & 5.20 \\
\addlinespace
12 & 2 & 1000 & 400 & 2.62 & 1000 & 400 & 2.72 \\
12 & 3 & 1000 & 400 & 2.80 & 1000 & 400 & 2.74 \\
12 & 5 & 1000 & 400 & 2.75 & 1000 & 400 & 2.75 \\
12 & 8 & 1000 & 400 & 2.63 & 1000 & 500 & 2.67 \\
12 & 16 & 1000 & 400 & 2.64 & 1000 & 400 & 2.71 \\
\addlinespace
16 & 2 & 1000 & 400$^{\dagger}$ & 1.60 & 1000 & 400 & 2.38 \\
16 & 3 & 1000 & 400$^{\dagger}$ & 2.27 & 1000 & 400 & 2.60 \\
16 & 5 & 1000 & 400$^{\dagger}$ & 2.17 & 1000 & 400 & 2.30 \\
16 & 8 & 1000 & 400 & 2.44 & 1000 & 400 & 2.28 \\
16 & 16 & 1000 & 400 & 2.37 & 1000 & 400 & 2.63 \\
\bottomrule
\end{tabular}
\end{table}

\paragraph{Escape is monotone in $R_{\text{old}}$, and the gate saturates where the
loss derivative does not.} On the 1000-step gate, row means are 6000, 2400, 1000,
1000, 1000 in seed 0, 6600, 3400, 1600, 1000, 1000 in seed 1 and 4800, 2600, 1000,
1000, 1000 in seed 2, so the gate saturates at $R_{\text{old}} \geq 8$ in seeds 0 and 2,
where 17 and 15 of 25 cells escape at the first evaluation. Seed 2's peak row means are
4220, 1960, 820, 400 and 375, the last two excluding $(12,8)$, $(12,16)$ and $(16,5)$,
where the derivative rises monotonically into the cosine ramp and the search returns a
position inside the decay rather than at formation. \S\ref{sec:escape} lists four seed-2
runs whose peak height is exceeded by the decay; the fourth, $(16,16)$, has its peak at
400 steps and is retained here, only its height being affected. Four hundred steps is the smallest
peak among seeds 0 and 1 rather than a grid limit: the derivative is computable from step
200, and seed 2 attains 300 at $R_{\text{old}}=16$, $\Delta D=3$.

\begin{figure}[t]
\centering
\includegraphics[width=0.88\textwidth]{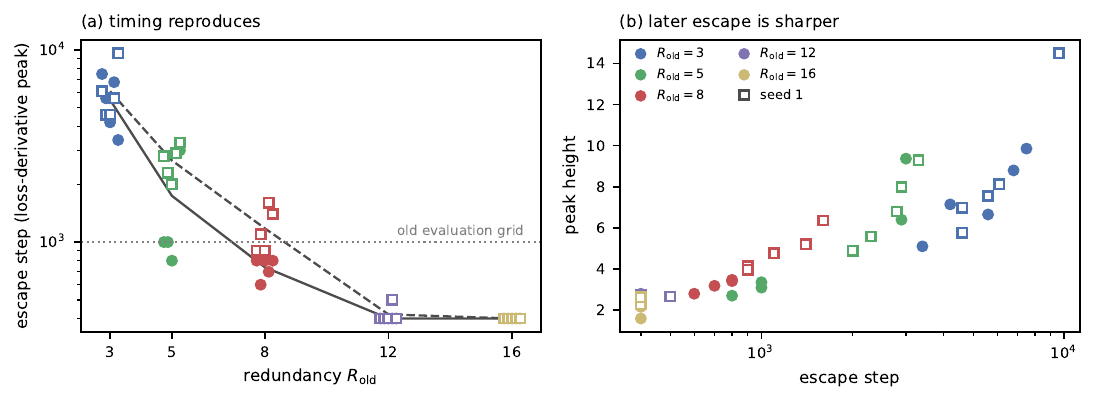}
\caption{Escape located from the peak of $d\,\mathrm{loss}/d\log\mathrm{step}$,
recorded every 100 steps with no threshold; filled markers seed 0, open markers seed 1.
Left: peak position against $R_{\mathrm{old}}$, with the 1000-step evaluation grid
marked. Right: peak height against peak position over all 50 runs.}
\label{fig:traj}
\end{figure}

A note on the escape threshold. We use 0.95 on the copy diagnostic; the
transitions are sharp enough that the choice does not matter within the analyzed
grid --- the diagnostic moves from below 0.05 to above 0.99 between consecutive
evaluations in every cell, so any threshold in $[0.05, 0.99]$ gives the same
escape step. It does matter in one excluded cell: at $R_{\mathrm{old}}=2$,
$\Delta D = 16$ the diagnostic reads 0.839 at step 4000 and 0.998 at 5000, so a
0.5 threshold would report escape at 4000 and our 0.95 threshold reports 5000. We
use 5000 in App.~\ref{app:posneg} and note the ambiguity there.

\section{Budget sensitivity and post-formation drift}
\label{app:drift}

Every readout in this paper is taken at a fixed budget, so it matters how much
of the map is a property of that choice. We ran two cells at three budgets each
and tracked one of them across 13 post-formation checkpoints.

\paragraph{The interior is stable; the boundary is not.} At $R_{\mathrm{old}}=16$, $\Delta D = 2$ the sign fraction is 0.995--1.000 across
an $8\times$ budget range while the median varies between $+1.81$ and $+2.43$
without ordering by budget --- the spread expected from mass compression at this
effect size (\S\ref{sec:degrade}), while the sign fraction is
effectively saturated at both ends. At $R_{\mathrm{old}}=3$, $\Delta D = 2$ ---
on the boundary, where \S\ref{sec:flip} declines to assign a phase --- the sign
fraction moves by 0.55 over the same kind of range.

\begin{table}[h]
\centering
\caption{Readout against step budget. The first two cells vary the budget at fixed
seed; the last three hold the escape step fixed at 1000 under both budgets, so any
difference between their rows is in the schedule rather than in when the circuit
formed. Medians are gated at mass $\geq 0.5$. The 12000-step row is from the earlier round of this appendix and is resolved on a finer evaluation grid, hence its non-multiple-of-1000 escape step.}
\label{tab:budget}
\begin{tabular}{llrrrl}
\toprule
cell & budget & escape & median & sign fraction & seed \\
\midrule
$R_{\mathrm{old}}=16$, $\Delta D=2$ & 4000  & 1000 & $+2.363$ & 0.995 & 0 \\
                                    & 16000 & 1000 & $+1.811$ & 0.998 & 0 \\
                                    & 32000 & 2000 & $+2.433$ & 1.000 & 0 \\
\midrule
$R_{\mathrm{old}}=3$, $\Delta D=2$  & 12000 & 2508 & $-0.000$ & 0.43 & 0 \\
                                    & 16000 & 8000 & $-0.011$ & 0.354 & 0 \\
                                    & 32000 & 6000 & $+0.262$ & 0.907 & 0 \\
\midrule
$R_{\mathrm{old}}=8$, $\Delta D=2$  & 16000 & 1000 & $-0.006$ & 0.45 & 1 \\
                                    & 32000 & 1000 & $+0.338$ & 0.98 & 1 \\
\midrule
$R_{\mathrm{old}}=16$, $\Delta D=16$ & 16000 & 1000 & $-0.009$ & 0.47 & 1 \\
                                     & 32000 & 1000 & $+0.102$ & 0.79 & 1 \\
\midrule
$R_{\mathrm{old}}=16$, $\Delta D=5$ & 16000 & 1000 & $+0.040$ & 0.58 & 1 \\
                                    & 32000 & 1000 & $+0.013$ & 0.54 & 1 \\
\bottomrule
\end{tabular}
\end{table}

The rows differ in what changes. In the first cell escape happens at the first or
second evaluation under every budget, so all three readouts are taken long after the
circuit exists and they agree. In the second the escape step itself moves with the
schedule (\S\ref{sec:limits}), so the three readouts are taken at different distances
past formation and they do not agree.

In the last three rows the readouts are taken at the same distance past formation, so
any difference is in the trajectory rather than in when it started. Two move and one
does not, and the two that differ most are the same row and the same seed. Doubling the
cosine period is therefore not a uniform sharpening: it changes where some runs stop
along the flat direction of \S\ref{sec:flat} and leaves others where they were. This is the basis for reporting the
location of the boundary as budget-dependent while treating the existence of the two
regions as not.

\paragraph{Post-formation trajectories drift, and in one run the drift crosses
zero.} Across the 13 evaluations from step 8000 to 32000 in the
$R_{\mathrm{old}}=3$, $\Delta D = 2$ run at the 32000-step budget, the mean of
$\Delta$ has standard deviation 0.49 nats over a range $[+1.33, +3.15]$, while the
sign fraction has standard deviation 0.03 over $[0.85, 0.95]$. Accuracy is 1.000
and the copy diagnostic is 1.000 at every one of those points, so nothing in
behavior distinguishes them. This is the second of the three instabilities that
motivate reporting the sign fraction rather than the mean (\S\ref{sec:dv}); the
third is below.

That run is not typical of the grid. Over the 69 runs with at least five post-formation
probe points, the sign fraction has within-run standard deviation between 0.006 and 0.303
with a median of 0.069 --- more than twice the value in the cell above. The largest is the
$R_{\mathrm{old}}=3$, $\Delta D=3$ run of seed 1, where the sign itself crosses 7000 steps
after the circuit forms (\S\ref{sec:plateau}). Per-seed medians are 0.069, 0.079 and
0.063, so the term does not shrink with a third seed, and the depth arm of App.~\ref{app:depth} spans 0.007--0.313 over its twelve runs, with
the largest at four layers, so the term does not shrink at other depths either.

\paragraph{The mass gate moves more at long budgets, and why.} Gating per
document on contrast-pair mass $\geq 0.5$ shifts the cell median by 1\% at 4000
steps and 16\% at 32000 in the $R_{\mathrm{old}}=16$, $\Delta D = 2$ cell. The
cause is visible in the per-document distribution rather than in any summary.
At 4000 steps, 2 of 400 documents have $\Delta > 8$ nats and the largest is
10.86. At 32000 a growing subpopulation exceeds 8 nats and the distribution between
8 and 14 is close to flat rather than decaying, with a maximum of 14.26 against the
analytic ceiling $2\kappa = 16$. The bulk of the distribution is essentially
unchanged --- hence a median shift of 16\% against a much larger change in the tail
--- so what the longer budget produces is a subpopulation of documents on which the
model has become nearly saturated, not a uniform sharpening.

\paragraph{A cross-budget consistency check on the first round.} An earlier
round of the grid at 12000 steps overlaps the main grid in six cells. Sign
fractions agree to within 0.02 in the four interior cells ($R_{\mathrm{old}}=16$ at
$\Delta D \in \{2,5,16\}$: 1.00/1.00, 1.00/1.00, 0.98/0.96) and move in the two boundary
cells ($R_{\mathrm{old}}=3$, $\Delta D = 2$: 0.43/0.354; $R_{\mathrm{old}}=5$,
$\Delta D = 5$: 0.69/0.49). The pattern matches the
one above --- interior cells reproduce across budgets, cells near the boundary
move --- and it is obtained at no additional cost, since those runs exist for
other reasons. The two rounds differ in step count only; we do not pool them.

\subsection{The cross-seed range does not narrow where it is widest}

If the spread of \S\ref{sec:flip} were the residue of incomplete convergence, the range
across seeds at a fixed cell should narrow as training continues past formation. We
computed it at every probe step common to all three seeds, restricted to steps
8000--16000 so that every cell is compared over the same absolute interval and the same
stretch of the schedule. Twenty of 25 cells have four or more such points; the other five
are at $R_{\mathrm{old}}=3$, $\Delta D \in \{2,16\}$, whose escape steps of 8000 and 10000
leave one point, and at $R_{\mathrm{old}}=16$, $\Delta D \in \{2,5,16\}$, where one seed
has two.

Averaged over cells the range falls from 0.335 to 0.305 between the first and second half
of that window, which is not significant: 13 of 20 cells narrow at $p = 0.077$ on a
one-sided Wilcoxon test. In the three widest cells it does not narrow at all. At
$R_{\mathrm{old}}=3$, $\Delta D=8$ the mean range is 0.882 early and 0.884 late; at $3, 5$
it is 0.949 and 0.899, both above the terminal 0.845; and at $3, 3$ it widens from 0.491 to
0.537, that being the window in which the seed-1 run's sign fraction crosses
(\S\ref{sec:plateau}).

Where the range does narrow, the bound explains it. The mean sign fraction across seeds
correlates with the ratio of late to early range at Spearman $\rho = -0.70$ over the 20
cells: the range collapses from 0.038 to 0.012 at $R_{\mathrm{old}}=12$, $\Delta D=5$ and
from 0.040 to 0.012 at $12, 8$, where all three seeds sit above 0.99 on a quantity confined
to $[0,1]$ and no range wider than 0.02 is available to them. The three widest cells sit at
0.45 to 0.74, where a range near 0.9 is available and the observed range is within 8\% of
it. Two cells at intermediate level widen rather than narrow ($5, 5$ at 0.69 and $8, 16$ at
0.76), so the pattern is not a gradient in either axis but a consequence of how much room
each cell has.

What this supports is that the cells carrying the paper's largest cross-seed differences
are not observably converging over the checkpoints available. It does not establish that
they would fail to converge under an arbitrarily longer schedule, which
\S\ref{sec:limits} already declines to claim.

\section{Breaking the collinearities}
\label{app:collin}

Two covariates of \S\ref{sec:limits} are collinear with the axes by construction: slot
count with $R_{\mathrm{old}}$, and the positional ceiling with $\Delta D$. Three arms below
hold one of them fixed and rerun. The first two both match slot count and differ in what
they cannot hold fixed alongside it --- document length in one, copy dispersion in the
other --- which is what lets them be read against each other.

\subsection{Slot count}
\label{app:slotmatch}

Slot count falls monotonically in $R_{\mathrm{old}}$ (Table~\ref{tab:covar}) and
cannot be held constant at fixed length and $p_{\mathrm{update}}$
(\S\ref{sec:limits}). It can be matched across two cells by shortening the
documents in one. Calibrating on 2000 documents, $n_{\mathrm{stmts}} \sim U[23,33]$
brings $R_{\mathrm{old}}=8$, $\Delta D=2$ to 8.95 slots against 8.73 for
$R_{\mathrm{old}}=16$, and brings copy dispersion most of the way with it: the two cells
differ by 11\% in dispersion where the main grid has them 41\% apart
(Table~\ref{tab:pupdcal}, where all four figures are measured on one path). Slot count is
matched and dispersion is nearly matched; realized redundancy is not, and moves counter to
the change. Training and evaluation are otherwise identical to the main grid.

\begin{table}[h]
\centering
\caption{Slot-matched arm. The shortened cell has the slot count of $R_{\mathrm{old}}=16$
and most of its copy dispersion (Table~\ref{tab:pupdcal}) while retaining
$R_{\mathrm{old}}=8$. The median
column is gated at contrast-pair mass $\geq 0.5$; medHi and medLo split all documents
with edit domain at each cell's own $q_{\mathrm{kept}}$ median and are ungated, so they
are not directly comparable to the median column where mass is low.
The two reference rows carry no split here (---); their within-cell splits are in
Table~\ref{tab:within}, where the seed-1 row at $R_{\mathrm{old}}=8$ is one of the six
cells whose pooled median falls below the 0.15-nat floor of \S\ref{sec:covar}. The seed-1 reference row is itself one of the 13 non-replicating cells ($+0.113$ at 0.74, against $+1.811$ at 1.00 in seed 0),
so percentages in the text are computed against the seed-0 reference and
baseline and the seed-1 and seed-2 rows are read as directional replication.}
\label{tab:slotmatch}
\begin{tabular}{lrrrlrlrr}
\toprule
cell & slots & $q_{\mathrm{kept}}$ & tokens & median & frac$+$ & $p$
& medHi & medLo \\
\midrule
$R_{\mathrm{old}}=8$ main, s0      & 14.59 & 5.75 & 204 & $+0.037$ & 0.66 & 9e-11  & $+0.086$ & $+0.003$ \\
$R_{\mathrm{old}}=8$ main, s1      & 14.59 & 5.75 & 204 & $-0.005$ & 0.45 & 0.06   & --- & --- \\
\midrule
$R_{\mathrm{old}}=8$ shortened, s0 &  8.95 & 4.02 & 116 & $+0.337$ & 0.95 & 2e-81 & $+0.413$ & $+0.172$ \\
$R_{\mathrm{old}}=8$ shortened, s1 &  8.95 & 4.02 & 116 & $+0.821$ & 0.99 & 4e-104 & $+1.070$ & $+0.432$ \\
$R_{\mathrm{old}}=8$ shortened, s2 &  8.95 & 4.02 & 116 & $+1.351$ & 0.98 & 3e-96 & $+1.683$ & $+1.109$ \\
\midrule
$R_{\mathrm{old}}=16$ main, s0     &  8.73 & 8.94 & 204 & $+1.811$ & 1.00 & 3e-118 & $+2.274$ & $+1.351$ \\
$R_{\text{old}} = 16$ main, s1 & 8.73 & 8.94 & 204 & $+0.113$ & 0.74 & 2e-21 & --- & --- \\
\bottomrule
\end{tabular}
\end{table}

Matching slot count moves the readout toward $R_{\mathrm{old}}=16$ on both summaries,
but only one of them reproduces across seeds. The sign fraction covers 85\%, 97\% and
95\% of the gap in seeds 0, 1 and 2 (0.95, 0.989, 0.983 against a reference of 1.00
and a baseline of 0.66), and all three lie outside the three-seed range of the unshortened cell (0.450--0.763), so the arm clears the replication criterion of
\S\ref{sec:flip}. One consequence of that criterion applies to the arm itself. In seed 1 the
shortened cell reads 0.989 against a reference of 0.736 in the same seed, so
matching slot count overshoots the reference rather than approaching it. The
percentages above are computed against the seed-0 reference, whose own
cross-seed spread (0.998 to 0.736) is of the same order as the gap being
closed: the arm establishes a direction, and the quantity it would have to be
measured against is not stable enough across seeds to support a fraction. The median does not: it covers 17\%, 44\% and 74\% of the same gap
($+0.337$, $+0.821$, $+1.351$ against $+1.811$), a fourfold spread across seeds that
brackets the effect we are trying to attribute. The advantage $R_{\mathrm{old}}$
retains in magnitude is 5.4-fold in seed 0 and 1.34-fold in seed 2, so we do not
report it. Read jointly: the length--sparsity complex accounts for the \emph{direction} at this point
in the grid; whether it also accounts for the \emph{magnitude} is not resolved by three
seeds, and the single-seed version of this arm would have said otherwise. Which component
of that complex carries the direction is also unresolved: App.~\ref{app:pupd} matches slot
count at fixed length instead and does not settle it.

This arm is the paper's clearest instance of the asymmetry \S\ref{sec:dv} relies on. Over
three seeds of one configuration the sign fraction spans 0.039 and the median spans
1.014, a factor of 26 difference in relative spread, with accuracy 1.000 and the copy
diagnostic 1.000 in all three.

\paragraph{The remaining confound runs the wrong way.} Shortening also lowers
realized redundancy, from 5.75 to 4.02 --- away from the 8.94 of the reference
cell, not toward it. Since the readout scales with realized redundancy both across
and within cells (\S\ref{sec:covar}), this covariate predicts a \emph{decrease}. We
observe an increase, so the change cannot be attributed to it. The within-cell
split sharpens this: the shortened cell's low-redundancy half reads $+0.172$,
twice the main-grid cell's high-redundancy half at $+0.086$, on documents carrying
fewer surviving copies. Two subsets ordered oppositely in redundancy are ordered
the same way as the cells. This holds in all three seeds and strengthens with each:
the shortened cell's low-redundancy half reads $+0.172$, $+0.432$ and $+1.109$ across
seeds 0, 1 and 2, against $+0.086$ for the main cell's high-redundancy half.

\paragraph{What this arm does not establish.} Shortening changes document length along with
slot count (204 tokens to 116), and length governs attention span and the effective range
of the positional encoding. App.~\ref{app:pupd} matches slot count at fixed length instead,
by moving $p_{\mathrm{update}}$; that arm holds length to within a token but removes almost none of the dispersion difference, so between the two arms slot count is never matched with
both of its companions held. What this arm establishes on its own is that the effect is not
carried by realized redundancy, which the confound above settles independently of either
separation. Read against \S\ref{sec:limits}: one confounded reading has become two readings
confounded in different ways, which is weaker than control and stronger than a single
measurement.

\subsection{Slot count at fixed length}
\label{app:pupd}

Raising $p_{\mathrm{update}}$ makes each filler slot absorb more statements, so at fixed
statement count it lowers the number of distinct slots without touching document length.
Slot count can therefore be matched across two cells by moving that parameter instead of
shortening the documents, which is the separation App.~\ref{app:slotmatch} cannot perform.
We ran it at $\Delta D = 2$, three seeds per leg, with $p_{\mathrm{update}} = 0.90$ at
$R_{\mathrm{old}}=8$ and $0.28$ at $R_{\mathrm{old}}=16$ against 0.5 in the main grid.
Calibration on 2000 documents (Table~\ref{tab:pupdcal}) gives 11.51 slots against 11.55, a
match to 0.04 slots, at 205 tokens on both legs; the main-grid rows reproduce
Table~\ref{tab:slotmatch} to 0.04 slots and 0.10 in realized redundancy. Training and
evaluation are otherwise identical to the main grid, and the two legs differ from it in
$p_{\mathrm{update}}$ alone.

\begin{table}[h]
\centering
\small
\caption{Calibration of the $p_{\mathrm{update}}$ arm, 2000 documents per configuration.
\emph{q\_gap} is the generator's per-document antecedent distance for the queried slot, the
same quantity as \emph{ant}$_q$ in Table~\ref{tab:inv}; the two measurement paths read
0.4--0.5 apart on the cells they share, so dispersion is compared within a path throughout.
\emph{q\_kept} is realized redundancy, \emph{tail0} the fraction of documents with no update
after the queried rebinding, and $\rho = p_{\mathrm{update}} / (1 + p_{\mathrm{update}}
R_{\mathrm{old}})$ the filler-side update density. The shortened row is the arm of
App.~\ref{app:slotmatch}, remeasured here so that all four dispersion figures come from one
path; it reproduces that table's slot count, realized redundancy and token count to within
1.1\% and its dispersion figure does not.}
\label{tab:pupdcal}
\begin{tabular}{lrrrrrrr}
\toprule
cell & $p_{\mathrm{update}}$ & slots & q\_gap & q\_kept & tokens & tail0 & $\rho$ \\
\midrule
$R_{\mathrm{old}}=8$ & 0.28 & 18.87 & 7.60 & 5.69 & 205.3 & 0.840 & 0.086 \\
\quad main           & 0.50 & 14.60 & 7.37 & 5.84 & 204.4 & 0.822 & 0.100 \\
\quad arm            & 0.90 & 11.51 & 7.48 & 5.94 & 205.2 & 0.791 & 0.110 \\
\quad shortened      & 0.50 &  8.94 & 5.82 & 3.98 & 117.3 & 0.821 & 0.100 \\
\addlinespace
$R_{\mathrm{old}}=16$ arm & 0.28 & 11.55 & 5.51 & 8.90 & 205.2 & 0.900 & 0.051 \\
\quad main                & 0.50 &  8.69 & 5.23 & 9.04 & 205.0 & 0.898 & 0.056 \\
                          & 0.90 &  6.75 & 5.23 & 9.43 & 204.9 & 0.896 & 0.058 \\
\bottomrule
\end{tabular}
\end{table}

\begin{table}[h]
\centering
\small
\caption{Readouts in the $p_{\mathrm{update}}$ arm. The median column is gated at
contrast-pair mass $\geq 0.5$; medHi and medLo split all documents with edit domain at each
cell's own realized-redundancy median and are ungated, so they are not comparable to the
median column where mass is low. Reference rows give the main grid's sign fraction only
(Table~\ref{tab:spread}); their medians and controls are in Table~\ref{tab:stats} and
Table~\ref{tab:within}. The seed-0 row of the $R_{\mathrm{old}}=8$ leg carries a control of
$-0.020$ against its median of $-0.055$ and is not interpretable; its sign fraction is
bracketed.}
\label{tab:pupd}
\begin{tabular}{llrrrrrr}
\toprule
cell & $p_{\mathrm{update}}$ & seed & median & frac$+$ & mass & medHi & medLo \\
\midrule
$R_{\mathrm{old}}=8$ main  & 0.50 & 0 & --- & 0.662 & --- & --- & --- \\
                           & 0.50 & 1 & --- & 0.450 & --- & --- & --- \\
                           & 0.50 & 2 & --- & 0.763 & --- & --- & --- \\
\midrule
$R_{\mathrm{old}}=8$ arm   & 0.90 & 0 & $-0.055$ & [0.154] & 1.00 & $-0.063$ & $-0.047$ \\
                           & 0.90 & 1 & $+2.852$ & 1.000 & 0.66 & $+4.371$ & $+3.540$ \\
                           & 0.90 & 2 & $+1.159$ & 0.997 & 0.97 & $+1.288$ & $+0.914$ \\
\midrule
$R_{\mathrm{old}}=16$ arm  & 0.28 & 0 & $+0.936$ & 0.997 & 0.99 & $+0.999$ & $+0.862$ \\
                           & 0.28 & 1 & $+1.235$ & 1.000 & 0.97 & $+1.441$ & $+1.060$ \\
                           & 0.28 & 2 & $+2.543$ & 1.000 & 0.82 & $+3.691$ & $+2.311$ \\
\midrule
$R_{\mathrm{old}}=16$ main & 0.50 & 0 & --- & 0.998 & --- & --- & --- \\
                           & 0.50 & 1 & --- & 0.736 & --- & --- & --- \\
                           & 0.50 & 2 & --- & 0.972 & --- & --- & --- \\
\bottomrule
\end{tabular}
\end{table}

\paragraph{The arm does not reproduce the shortened arm, and is not powered to refute it.}
If slot count carries the effect, two cells carrying 11.5 slots should read alike: the
readout at $R_{\mathrm{old}}=8$ should rise toward $R_{\mathrm{old}}=16$ and the readout at
$16$ should fall toward $8$. Both rise. The sign fraction goes to 0.154, 1.000 and 0.997 at
$R_{\mathrm{old}}=8$ against 0.662, 0.450 and 0.763 in the main grid, and to 0.997, 1.000
and 1.000 at $16$ against 0.998, 0.736 and 0.972. Two facts prevent us from calling this a
refutation. Averaging the sign fraction over seeds leaves the difference between the two
cells where it was, 0.277 in the main grid against 0.282 here, while taking the median over
seeds collapses it from 0.310 to 0.003, because the median discards the one seed that
reverses; a conclusion that inverts with the pooling statistic is not a conclusion. And the
reference contrast is itself the size of its own noise: the main-grid difference between
these cells is 0.277 in sign fraction against cross-seed ranges of 0.313 and 0.261
(Table~\ref{tab:spread}). The shortened arm covers 85--97\% of the same difference with no
seed reversing; this arm does not, and three seeds at this signal-to-noise ratio cannot
distinguish a failure to reproduce from a failure to detect.

\paragraph{Nor can it isolate slot count, for a reason independent of power.} Copy
dispersion is a nearest-neighbour distance among the queried slot's surviving copies, and its
product with realized redundancy is an empirical invariant of each row under this parameter:
across $p_{\mathrm{update}} \in [0.28, 0.90]$ that product is constant to within 4\% (43.0 to
44.4 at $R_{\mathrm{old}}=8$, 47.3 to 49.3 at $16$) and dispersion itself moves by under 6\%
(7.37 to 7.60 and 5.23 to 5.51) while slot count changes by a factor of 1.64 and 1.71
(Table~\ref{tab:pupdcal}). The parameter acts only on filler slots --- the queried slot is
built with its update forced, so its own copy count does not read $p_{\mathrm{update}}$ ---
and it therefore moves slot count without moving the geometry dispersion is computed from.
Matching slot count by this route leaves the two cells 36\% apart in dispersion where the
main grid has them 41\% apart, while shortening brings the same difference to 11\%. Read that
against the results: the effect appears in the arm that removes most of the dispersion
difference and moves length, and is not detectable in the arm that holds length and removes
almost none of it. Under the mean pooling above, that is the pattern an account in which copy
dispersion rather than slot count carries the readout would predict, and the same account
fits the main grid, where dispersion falls from 13.32 at $R_{\mathrm{old}}=3$ to 4.71 at $16$
(Table~\ref{tab:inv}) as the readout rises. Under median pooling it does not hold, and the
ordering rests on two arms, so the two of them point at such an account without establishing
it.

\paragraph{What survives either pooling.} Slot count falls 21\% at $R_{\mathrm{old}}=8$ and
rises 33\% at $16$ while the readout rises in both legs, and the filler update density
$\rho$ also moves in opposite directions (0.100 to 0.110 against 0.056 to 0.051) while the
readout does not. No monotone function of either quantity describes the two legs together.
This is a statement about the signs of four differences and does not depend on how seeds are
pooled; it is also weaker than it sounds, since it excludes monotone dependence on these
covariates alone rather than identifying what the readout does follow.

\paragraph{One readout is not interpretable.} At $R_{\mathrm{old}}=8$, seed 0 the median is
$-0.055$ and the filler-slot control reads $-0.020$, 36\% of it with the two signs agreeing.
The control is outside the $-0.009$ to $+0.015$ range the grid spans, and the argument
App.~\ref{app:oracle} uses for low-median cells --- that the control is opposite in sign
to the readout and at most 5\% of it --- is unavailable here. We
therefore do not read this run's 0.154 as a reversal, and it is the seed the median pooling
above discards. In the $R_{\mathrm{old}}=16$ leg, where the arm's conclusion would otherwise
rest, the three controls are $-0.001$, $+0.002$ and $-0.006$ against ungated medians of
$+0.938$, $+1.258$ and $+2.895$ (gated: $+0.936$, $+1.235$, $+2.543$, as in
Table~\ref{tab:pupd}).

\paragraph{Four findings that replicate.} All six runs reach in-distribution accuracy 1.000,
the copy diagnostic 1.000 and the \textsc{retrieval} state, with the gate crossed at step
1000 in five and 2000 in one, so a configuration absent from the main grid --- at
$p_{\mathrm{update}} = 0.90$ nine filler slots in ten receive an update --- does not disturb
circuit formation. The within-cell gradient of \S\ref{sec:covar} runs in the direction of
each cell's own effect in all six, including the negative one ($-0.063$ against $-0.047$).
Contrast-pair mass is strictly decreasing in the size of the effect across all six runs
(0.999, 0.989, 0.973, 0.971, 0.821, 0.664 against ungated medians from $-0.055$ to
$+4.126$), a rank correlation of $-1.000$ and the sharpest instance in the paper of
\S\ref{sec:degrade}. And the cross-seed ranges follow the bound of App.~\ref{app:drift}:
0.003 at $R_{\mathrm{old}}=16$, where all three seeds sit above 0.997 and no larger range is
available, against 0.846 at $8$, where the pooled level of 0.72 leaves room for about 0.9.
The first of those two is a ceiling effect and not evidence of stability.

\paragraph{No single knob isolates slot count.} Slot count responds to three parameters and
each moves something else: $n_{\mathrm{stmts}}$ moves length, $p_{\mathrm{update}}$ moves the
filler-side update density, and $\mathrm{spread}$ moves the window the queried slot's copies
are placed in and hence both dispersion and how many of them survive. Because
$p_{\mathrm{update}}$ \emph{is} the filler-side update probability, an account in which the
readout responds to slot sparsity and one in which it responds to filler update density make
the same prediction in this arm as well. What the two arms buy jointly is one negative result
and one boundary: realized redundancy does not carry the effect
(App.~\ref{app:slotmatch}, where it moves counter to it and the within-cell split confirms
the direction), and neither arm decomposes the length--sparsity complex. A third arm moving
$\mathrm{spread}$ and $p_{\mathrm{update}}$ together might match slot count, length and
dispersion at once; whether three knobs can hit those three targets while realized redundancy
also responds to one of them we have not checked, and we did not run it. The two arms here
bracket the question rather than closing it.

\begin{table}[ht]
\caption{Within-cell dose-response, three seeds. Each cell is split at its own
median realized redundancy; medHi and medLo are the medians of the two halves,
over all documents with edit domain and ungated, so they are not comparable to
the gated medians of Table~\ref{tab:stats} where mass is low. The gradient runs
in the direction of the cell's own effect in 65 of the 75 runs. $^{a}$ marks the
six exceptions whose pooled median is below 0.15 nats in absolute value, where
there is no effect to stratify. $^{b}$ marks the four that invert a gradient
while carrying an effect: the inversion is 8\% and 3\% of the pooled median in
seeds 0 and 1, and 17\% and 10\% in the two contributed by seed 2. The two-seed
version of this table carried no inversion above 8\%, so the claim the third seed
removes is that inversions are confined to gradients negligible relative to their
cell; what survives is that the split tracks the size of the effect, at Spearman
$\rho = 0.865$ over the 50 runs of seeds 0 and 1.}
\label{tab:within}
\centering
\small
\begin{tabular}{ll rr rr rr}
\toprule
& & \multicolumn{2}{c}{seed 0} & \multicolumn{2}{c}{seed 1} & \multicolumn{2}{c}{seed 2} \\
\cmidrule(lr){3-4} \cmidrule(lr){5-6} \cmidrule(lr){7-8}
$R_{\text{old}}$ & $\Delta D$ & medHi & medLo & medHi & medLo & medHi & medLo \\
\midrule
3 & 2 & $-0.016$ & $-0.005$ & $-0.051$ & $-0.041$ & $+0.029$ & $+0.002$ \\
3 & 3 & $-0.046$\rlap{$^{a}$} & $-0.047$ & $+2.879$ & $+2.592$ & $+1.472$ & $+0.467$ \\
3 & 5 & $-0.120$ & $-0.069$ & $+7.458$ & $+0.430$ & $-0.017$ & $-0.011$ \\
3 & 8 & $-0.041$ & $-0.015$ & $-0.001$\rlap{$^{a}$} & $-0.001$ & $+1.486$ & $+0.870$ \\
3 & 16 & $-0.022$\rlap{$^{a}$} & $-0.024$ & $-0.032$ & $-0.017$ & $-0.032$ & $-0.007$ \\
\addlinespace
5 & 2 & $+0.083$ & $+0.018$ & $-0.043$ & $-0.013$ & $+0.481$ & $+0.322$ \\
5 & 3 & $+0.014$ & $+0.003$ & $+1.847$ & $+1.129$ & $+0.242$ & $+0.199$ \\
5 & 5 & $+0.013$ & $-0.009$ & $+0.059$ & $+0.006$ & $+0.300$ & $+0.175$ \\
5 & 8 & $+0.190$ & $+0.098$ & $+0.073$\rlap{$^{a}$} & $+0.136$ & $+0.596$ & $+0.158$ \\
5 & 16 & $+0.115$ & $+0.044$ & $+0.433$ & $+0.250$ & $+0.172$ & $+0.025$ \\
\addlinespace
8 & 2 & $+0.086$ & $+0.003$ & $+0.000$\rlap{$^{a}$} & $-0.019$ & $+0.113$ & $+0.081$ \\
8 & 3 & $+0.618$ & $+0.572$ & $+6.420$ & $+4.106$ & $+1.022$ & $+0.936$ \\
8 & 5 & $+0.476$ & $+0.250$ & $+2.326$ & $+1.752$ & $+0.997$ & $+0.974$ \\
8 & 8 & $+0.356$ & $+0.285$ & $+5.651$ & $+4.318$ & $+2.054$ & $+0.327$ \\
8 & 16 & $+0.295$ & $+0.142$ & $+0.507$ & $+0.287$ & $-0.016$ & $-0.008$ \\
\addlinespace
12 & 2 & $+3.747$ & $+3.136$ & $+7.589$ & $+2.943$ & $+0.833$ & $+0.318$ \\
12 & 3 & $+0.681$ & $+0.599$ & $+0.003$\rlap{$^{a}$} & $+0.023$ & $+4.017$\rlap{$^{b}$} & $+4.447$ \\
12 & 5 & $+6.119$ & $+3.987$ & $+1.042$\rlap{$^{b}$} & $+1.074$ & $+5.089$ & $+2.802$ \\
12 & 8 & $+3.164$ & $+2.098$ & $+2.301$ & $+1.076$ & $+1.987$ & $+0.610$ \\
12 & 16 & $+2.218$ & $+0.997$ & $+1.406$ & $+0.728$ & $+1.111$ & $+0.455$ \\
\addlinespace
16 & 2 & $+2.274$ & $+1.351$ & $+0.117$ & $+0.111$ & $+0.780$\rlap{$^{b}$} & $+0.923$ \\
16 & 3 & $+3.556$\rlap{$^{b}$} & $+3.857$ & $+1.129$ & $+0.543$ & $+1.448$ & $+0.608$ \\
16 & 5 & $+1.857$ & $+1.171$ & $+0.040$ & $+0.034$ & $+0.444$ & $+0.213$ \\
16 & 8 & $+2.344$ & $+1.967$ & $+0.099$ & $+0.048$ & $+1.420$ & $+0.945$ \\
16 & 16 & $+1.096$ & $+0.624$ & $-0.028$ & $+0.024$ & $+0.955$ & $+0.652$ \\
\addlinespace
\bottomrule
\end{tabular}
\end{table}

\subsection{Fixed-width bands}
\label{app:fixband}

The $\Delta D$ axis of the main grid varies band width with $d$, so the positional
ceiling $1/\lvert\mathrm{supp}(\Delta D)\rvert$ falls from 0.333 to 0.059 along it
(\S\ref{sec:limits}). We repeated the $R_{\mathrm{old}}=3$ row with
$\Delta D \sim U[d, d{+}8]$ at $d \in \{1,4,8,16\}$, three seeds each, holding the
ceiling at $1/9 = 0.111$ in every cell. Training and evaluation are otherwise identical
to the main grid. The $d=4$ cell coincides with the main grid's $R_{\mathrm{old}}=3,
\Delta D=8$ cell, whose band $[4,12]$ already has width 9; retrained under the new
configuration name it reproduces the original to all recorded digits in all three seeds,
which extends the bit-identical check of App.~\ref{app:gen} across configuration names.

\paragraph{The ceiling is held and the alias survives.} Measured positional collision is
0.120, 0.127, 0.121 and 0.131 against the analytic 0.111, a spread of 0.011 where the
main grid spans 0.069--0.353 (App.~\ref{app:cells}). All five generator invariants pass
in all four cells, and the antecedent-distance ratio that holds only partially in the
main grid improves: 1.19, 1.15, 1.09 and 0.98 against up to 1.22 there, so the
dispersion cue of \S\ref{sec:limits} weakens as the band widens. \textsc{recency} and
\textsc{rarity} collide at 1.000 in every cell, making Equation~\ref{eq:alias} a property
of the construction rather than of one band geometry, and \textsc{frequency} reaches
0.420 at $d=16$, extending the $1-\mathrm{domain}$ identity of App.~\ref{app:cells}
past the main grid's range ($0.420$ against $1-0.573$).

\paragraph{The cross-seed spread survives at constant ceiling.} Sign fractions are
0.870/0.104/0.204 at $d=1$, 0.098/0.477/0.977 at $4$, 0.319/0.359/0.418 at $8$ and
0.584/0.486 at $16$, where the third seed carries no readout (below). Three of the four
cells straddle 0.5, and row means by seed are 0.468, 0.357 and 0.533 against 0.270, 0.538
and 0.630 for the same row of the main grid. The $d=1$ cell spans 0.766 with every
endpoint individually decisive ($p = 1\times10^{-48}$ in the rarity-type direction,
$3\times10^{-55}$ and $3\times10^{-30}$ in the frequency-type one), so it is a fresh
instance of \S\ref{sec:flip} in a cell absent from the main grid. Because the ceiling is
identical across these four cells, the spread cannot be mediated by how much the
positional shortcut pays --- the one account of \S\ref{sec:flip} that the main grid,
where ceiling and $\Delta D$ are collinear, cannot exclude.

\paragraph{What this row does not settle.} The gradient \S\ref{sec:limits} names is at
$R_{\mathrm{old}}=16$ and this arm is at $3$, where the main grid has no $\Delta D$
ordering to begin with; the row means here (0.393, 0.517, 0.365, 0.535) have none either,
which is consistent with both accounts and distinguishes neither. Escape time also fails
to order: gate row means rise 5000, 5333, 6333 and 7667 steps with $d$, but the
loss-derivative peak at $d=16$ is 1400, 6500 and 10100 across the three seeds, a
sevenfold within-cell spread that swamps the between-cell differences. This is the
evidence behind reporting the $R_{\mathrm{old}}=3$ escape range rather than a $\Delta D$
trend (App.~\ref{app:states}).

\paragraph{One run acquires the circuit without solving the task.} At $d=16$, seed 2 ends
with the copy diagnostic at 1.000 and a loss-derivative peak of 10.16 at step 10100,
both signatures of circuit formation, but in-distribution accuracy of 0.128 --- $1.15$
times the positional ceiling --- and 0.05 on the no-tail-update stratum. Contrast-pair
mass is 0.12 with 3\% of documents clearing the per-document gate, so the cell carries no
readout and we exclude it. It is the only run we have seen in which the two conditions of
the \textsc{retrieval} label come apart: the copy diagnostic clears 0.95 while accuracy
does not clear 0.99, so the label falls to \textsc{neither} on a model that
demonstrably performs slot matching. This is a second and sharper form of
\S\ref{sec:plateau}'s point that gating on circuit formation is necessary but not
sufficient --- there the gate admits a run whose readout later reverses, here it would
admit a run that has not solved the task at all. Whether the circuit would convert to
accuracy past our budget or is a different solution we cannot say from one run.

\section{Summary statistics}
\label{app:stats}

\begin{figure}[t]
\centering
\includegraphics[width=\textwidth]{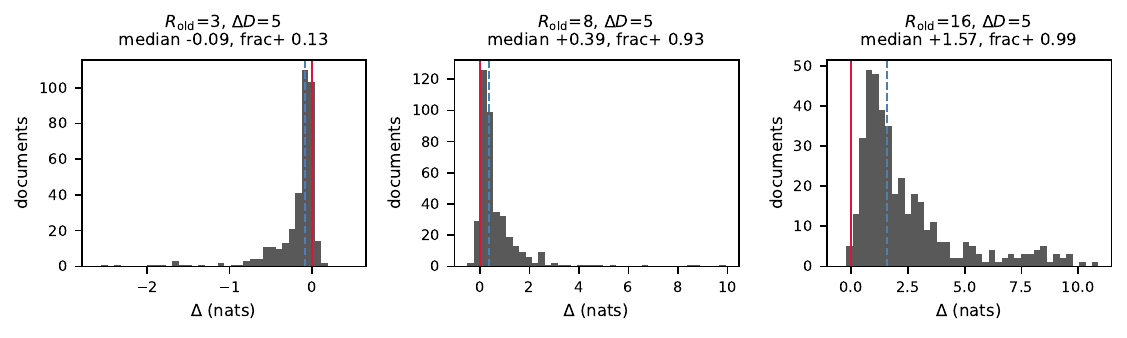}
\caption{Per-document readout in three cells along the $\Delta D = 5$ column.
Red line at zero, blue dashed at the median. Note the asymmetry in shape: the
negative-side distribution is concentrated near zero with a thin tail, while the
positive-side distributions are broad. Axes are not shared.}
\label{fig:dist}
\end{figure}

Table~\ref{tab:stats} reports every summary we computed for each cell of seed 0,
including
the ones \S\ref{sec:dv} argues against using. Seeds 1 and 2 enter through
Table~\ref{tab:spread}, which is the comparison the paper's central claim rests on;
its per-cell medians and controls are quoted in \S\ref{sec:sign} and
\S\ref{sec:degrade} where they bear on an argument. Medians and interquartile ranges are
over the documents on which the edit has a domain; the sign fraction is the
fraction of those documents on which $\Delta$ has the rarity-type sign, with
$p$ from a two-sided exact binomial test against 0.5. \emph{ctrl} is the median
of the magnitude-matched control edit on a non-queried slot. \emph{mass} is the
mean probability assigned to $\{v^\star, v_{\mathrm{truth}}\}$ under the edit and
\emph{mOK} the fraction of documents clearing the 0.5 gate.

\begin{table}[h]
\centering
\small
\caption{Per-cell readouts, main grid at 16000 steps, seed 0, 400 held-out
documents per cell. $n$ is the number with edit domain.}
\label{tab:stats}
\setlength{\tabcolsep}{2.8pt}
\begin{tabular}{rr rr rrr rr rr r rr}
\toprule
$R$ & $\Delta D$ & $n$ & yield & median & IQR low & IQR high & frac$+$ & $p$
& mean & trim & ctrl & mass & mOK \\
\midrule
3 & 2 & 364 & 0.91 & $-0.011$ & $-0.043$ & $+0.007$ & 0.35 & 3e-08 & $-0.109$ & $-0.034$ & $+0.000$ & 1.00 & 1.00 \\
3 & 3 & 362 & 0.91 & $-0.046$ & $-0.213$ & $+0.055$ & 0.36 & 3e-07 & $-0.170$ & $-0.115$ & $+0.001$ & 1.00 & 1.00 \\
3 & 5 & 364 & 0.91 & $-0.087$ & $-0.219$ & $-0.030$ & 0.13 & 4e-51 & $-0.218$ & $-0.150$ & $+0.004$ & 1.00 & 1.00 \\
3 & 8 & 328 & 0.82 & $-0.024$ & $-0.075$ & $-0.009$ & 0.10 & 1e-54 & $-0.073$ & $-0.055$ & $+0.000$ & 1.00 & 1.00 \\
3 & 16 & 274 & 0.69 & $-0.023$ & $-0.104$ & $+0.038$ & 0.41 & 2e-03 & $-0.094$ & $-0.058$ & $+0.001$ & 1.00 & 1.00 \\
\midrule
5 & 2 & 397 & 0.99 & $+0.066$ & $+0.008$ & $+0.137$ & 0.78 & 2e-30 & $+0.087$ & $+0.076$ & $-0.006$ & 1.00 & 1.00 \\
5 & 3 & 393 & 0.98 & $+0.010$ & $-0.013$ & $+0.037$ & 0.61 & 8e-06 & $+0.029$ & $+0.014$ & $+0.002$ & 1.00 & 1.00 \\
5 & 5 & 395 & 0.99 & $+0.003$ & $-0.033$ & $+0.049$ & 0.52 & 0.48 & $+0.176$ & $+0.014$ & $-0.001$ & 0.98 & 0.98 \\
5 & 8 & 380 & 0.95 & $+0.147$ & $+0.027$ & $+0.333$ & 0.80 & 2e-33 & $+0.293$ & $+0.199$ & $-0.000$ & 0.99 & 0.99 \\
5 & 16 & 343 & 0.86 & $+0.090$ & $+0.012$ & $+0.255$ & 0.80 & 2e-29 & $+0.298$ & $+0.154$ & $+0.004$ & 0.98 & 0.99 \\
\midrule
8 & 2 & 397 & 0.99 & $+0.037$ & $-0.015$ & $+0.159$ & 0.66 & 9e-11 & $+0.199$ & $+0.107$ & $-0.003$ & 1.00 & 1.00 \\
8 & 3 & 397 & 0.99 & $+0.604$ & $+0.350$ & $+1.080$ & 0.99 & 6e-113 & $+0.888$ & $+0.776$ & $-0.002$ & 0.99 & 1.00 \\
8 & 5 & 397 & 0.99 & $+0.385$ & $+0.156$ & $+0.872$ & 0.93 & 4e-78 & $+0.749$ & $+0.552$ & $+0.002$ & 0.98 & 0.98 \\
8 & 8 & 396 & 0.99 & $+0.338$ & $+0.191$ & $+0.626$ & 0.97 & 3e-97 & $+0.547$ & $+0.439$ & $+0.004$ & 0.99 & 0.99 \\
8 & 16 & 389 & 0.97 & $+0.243$ & $+0.106$ & $+0.532$ & 0.89 & 9e-62 & $+0.526$ & $+0.371$ & $+0.004$ & 0.99 & 0.99 \\
\midrule
12 & 2 & 400 & 1.00 & $+3.523$ & $+1.888$ & $+6.091$ & 1.00 & 8e-121 & $+4.258$ & $+4.067$ & $+0.001$ & 0.80 & 0.82 \\
12 & 3 & 399 & 1.00 & $+0.643$ & $+0.393$ & $+1.030$ & 0.99 & 2e-111 & $+1.024$ & $+0.791$ & $+0.001$ & 0.98 & 0.98 \\
12 & 5 & 398 & 0.99 & $+5.329$ & $+2.953$ & $+7.951$ & 1.00 & 1e-117 & $+5.527$ & $+5.465$ & $+0.014$ & 0.66 & 0.66 \\
12 & 8 & 398 & 0.99 & $+2.678$ & $+1.567$ & $+4.416$ & 1.00 & 3e-120 & $+3.391$ & $+3.178$ & $+0.010$ & 0.89 & 0.91 \\
12 & 16 & 396 & 0.99 & $+1.867$ & $+0.900$ & $+3.367$ & 0.98 & 2e-103 & $+2.468$ & $+2.239$ & $-0.001$ & 0.91 & 0.93 \\
\midrule
16 & 2 & 400 & 1.00 & $+1.827$ & $+1.095$ & $+3.153$ & 1.00 & 3e-118 & $+2.210$ & $+2.075$ & $+0.012$ & 0.98 & 0.99 \\
16 & 3 & 400 & 1.00 & $+3.625$ & $+2.088$ & $+5.923$ & 1.00 & 8e-121 & $+4.278$ & $+4.066$ & $-0.004$ & 0.82 & 0.83 \\
16 & 5 & 399 & 1.00 & $+1.575$ & $+0.935$ & $+3.057$ & 0.99 & 1e-115 & $+2.454$ & $+2.209$ & $+0.006$ & 0.91 & 0.91 \\
16 & 8 & 398 & 0.99 & $+2.216$ & $+1.538$ & $+3.240$ & 1.00 & 1e-117 & $+2.580$ & $+2.423$ & $+0.004$ & 0.95 & 0.98 \\
16 & 16 & 395 & 0.99 & $+0.891$ & $+0.364$ & $+2.011$ & 0.98 & 1e-101 & $+1.463$ & $+1.250$ & $+0.001$ & 0.98 & 0.98 \\
\bottomrule
\end{tabular}
\end{table}

\begin{table}[t]
\caption{Sign fraction per cell under three seeds at 16000 steps, 400 held-out
documents each, ordered by the range across seeds. Thirteen of 25 cells span more
than 0.3; the binomial standard error at $n=400$ is at most 0.025, so the largest
range is 35 standard errors. All 75 runs reach in-distribution accuracy
$\geq 0.999$ and terminate in the \textsc{retrieval} state, so no held-in
evaluation separates any pair of entries. \S\ref{sec:flip} treats a cell-to-cell difference as meaningful only when it exceeds the range in this table. Adding the
third seed widened the maximum from 0.845 to 0.879 and moved the cell carrying it
from $(3,5)$ to $(3,8)$.}
\label{tab:spread}
\centering
\small
\begin{tabular}{llrrrr@{\qquad}llrrrr}
\toprule
$R_{\mathrm{old}}$ & $\Delta D$ & s0 & s1 & s2 & range &
$R_{\mathrm{old}}$ & $\Delta D$ & s0 & s1 & s2 & range \\
\midrule
3  & 8  & 0.098 & 0.477 & 0.977 & 0.879 & 16  & 2  & 0.998 & 0.736 & 0.972 & 0.261 \\
3  & 5  & 0.126 & 0.972 & 0.270 & 0.845 &  3  & 16 & 0.405 & 0.167 & 0.349 & 0.238 \\
5  & 2  & 0.781 & 0.175 & 0.967 & 0.792 &  5  & 8  & 0.800 & 0.798 & 0.931 & 0.133 \\
3  & 3  & 0.365 & 0.934 & 0.930 & 0.569 &  8  & 5  & 0.932 & 1.000 & 0.947 & 0.068 \\
16 & 16 & 0.977 & 0.469 & 0.969 & 0.508 &  8  & 8  & 0.970 & 0.997 & 0.934 & 0.063 \\
3  & 2  & 0.354 & 0.138 & 0.625 & 0.487 &  5  & 16 & 0.796 & 0.844 & 0.784 & 0.060 \\
8  & 16 & 0.895 & 0.916 & 0.432 & 0.484 & 12  & 2  & 1.000 & 1.000 & 0.953 & 0.047 \\
12 & 3  & 0.990 & 0.570 & 0.998 & 0.428 & 16  & 3  & 1.000 & 0.965 & 0.985 & 0.035 \\
16 & 5  & 0.995 & 0.584 & 0.870 & 0.411 & 12  & 16 & 0.980 & 0.992 & 0.969 & 0.023 \\
5  & 3  & 0.613 & 0.997 & 0.964 & 0.384 &  8  & 3  & 0.992 & 0.997 & 0.985 & 0.012 \\
5  & 5  & 0.519 & 0.675 & 0.880 & 0.361 & 12  & 8  & 1.000 & 0.995 & 0.992 & 0.008 \\
16 & 8  & 0.997 & 0.670 & 0.992 & 0.327 & 12  & 5  & 0.997 & 0.995 & 1.000 & 0.005 \\
8  & 2  & 0.662 & 0.450 & 0.763 & 0.313 &     &    &       &       &       &       \\
\bottomrule
\end{tabular}
\end{table}

\paragraph{Mass and the gated median.} Contrast-pair mass is 1.00 with mOK 1.00 in all
five $R_{\mathrm{old}}=3$ cells of seed 0 and 0.98--1.00 at $R_{\mathrm{old}}=5$ and
$8$. It falls below 0.85 in ten cells across the three seeds, lowest at
$R_{\mathrm{old}}=12$, $\Delta D=2$ in seed 1 (0.59), at $12, 5$ in seed 0 and
$3, 5$ in seed 1 (both 0.66), and at $12, 5$ and $12, 3$ in seed 2 (0.69 and 0.70),
so mass tracks the realized effect rather than either axis. The $R_{\mathrm{old}}=12$ row of seed 0 shows the consequence for the median:
values run $+3.52$, $+0.64$, $+5.33$, $+2.68$, $+1.87$ in $\Delta D$ and the low cell
is the one with the highest mass (0.98 against 0.66 and 0.80 in its neighbours), so the
ordering tracks what the instrument can register rather than a mechanism gradient,
while the sign fraction is 0.98--1.00 across the row. The sign fraction is computed
over all documents with edit domain rather than over the gated subset, so it is
unaffected by the gate by construction (\S\ref{sec:dv}). The median column of
Table~\ref{tab:stats} is ungated; gated medians are quoted in the main text where an
argument depends on them (\S\ref{sec:sign}, \S\ref{sec:degrade}).

\paragraph{Mean against sign fraction.} The mean and the sign fraction disagree
in ordering at two places in the grid. At $R_{\mathrm{old}}=5$, $\Delta D=5$ the
mean is $+0.176$ --- larger than the mean at $\Delta D=3$ in the same row
($+0.029$) --- while the sign fraction is 0.52 against 0.61 and the binomial test
returns $p=0.48$. The mean is carried by four documents above $+8$ nats in a
distribution otherwise concentrated within $\pm 0.5$. At $R_{\mathrm{old}}=12$,
$\Delta D=3$ the mean is $+1.02$ against $+5.53$ two columns over, a fivefold
difference where the sign fraction differs by 0.01. Both are cells where the
5\%-trimmed mean sits closer to the median than to the mean, which is the
signature of a heavy tail rather than a shifted centre.

\paragraph{Distribution shape by sign.} Figure~\ref{fig:dist} shows the
per-document readout in three cells along the $\Delta D = 5$ column. The two
directions differ in shape as well as in sign: at $R_{\mathrm{old}}=3$ the
distribution is concentrated within $\pm 0.1$ nats with a thin negative tail
reaching $-3.4$, while at $R_{\mathrm{old}}=16$ it is broad and spans 0 to 12
nats. The same edit produces near-indifference plus a few strong reactions on one
side of the axis and a large shift in most documents on the other. This is
consistent with the low-$R_{\mathrm{old}}$ cells lacking a stable mechanism
rather than implementing an opposite one, the reading \S\ref{sec:flip} declines to
settle.

All five $R_{\mathrm{old}}=3$ cells are unimodal under this binning, so the
negative median is not a mixture of two populations using different rules. Two
cells show a second local maximum in the histogram
($\Delta D \in \{5, 8\}$), both adjacent to the mode and within counting noise of
it.

\section{Attempting to measure the geometry}
\label{app:flat}

\S\ref{sec:flat} argues from where the variance lives. We also tried to measure the
geometry directly. The measurement does not support the conclusion, and we report it
because the reason it fails applies to anyone attempting the same test.

\paragraph{What is computed.} At $R_{\mathrm{old}}=3, \Delta D=5$ and
$R_{\mathrm{old}}=16, \Delta D=2$, seed 0, at twelve checkpoints each, we take
$g_L = \nabla_\theta L$ on 512 fresh documents from the training distribution and
$g_\Delta = \nabla_\theta \mathbb{E}[\Delta]$ on the held-out probe set, form
$\hat u_{\Delta\perp}$ by removing the $g_L$ component from $g_\Delta$ and normalizing,
and evaluate $L$ and $\mathbb{E}[\Delta]$ at $\theta \pm \varepsilon \hat u$ at
$\varepsilon = 3\times10^{-4}\lVert\theta\rVert$, the one step size that is both above the
fp32 noise floor on $\Delta L$ and inside the linear regime. Every evaluation reuses the
same documents and the same edit pairs; all arithmetic is fp32 with TF32 disabled. Because
$\Delta$ is a difference of two log-odds sharing a normalizer, it reduces to a difference
of raw logits and the gradient does not pass through a softmax. Both runs are the
main-grid runs re-executed with periodic checkpointing; the first 2000 logged losses are
bit-identical to the originals, so checkpointing does not perturb the trajectory
(\S\ref{sec:inv}).

\paragraph{The readout reproduces.} At 16000 steps the median and sign fraction agree
with Table~\ref{tab:stats} and Table~\ref{tab:spread} to two decimals in both cells
($-0.0874$ / $0.126$ and $+1.8218$ / $0.998$). This is an independent check on the readout
arithmetic: the code path shares the edit construction of \S\ref{sec:probe} but computes
$\Delta$ from raw logits with no caching and no per-document loop.

\paragraph{Seven other readouts, built the same way.} The comparison the measurement needs
is against a direction constructed identically but trading a pair the objective
distinguishes. We built seven. Five are further edits from the probe suite, each moving one
candidate rule's prediction while leaving the ground truth fixed, with token count, answer
position, statement count and antecedent distance verified per document as in
App.~\ref{app:oracle}. \emph{Frequency}: the superseded value's copies are reduced to one
and the surplus rewritten as fresh singleton slots, so the majority value changes while
the most recent does not. \emph{Primacy}: every copy of the superseded value is relabelled
to one fresh value, changing which value occupies the slot's first position without
changing any count. \emph{Position}: the final queried statement is exchanged with an
earlier filler and the latest superseded copy is moved by the same distance, so $\Delta D$
grows while the distance to the nearest same-slot antecedent is preserved. \emph{Global
last update} contributes two edits with complementary domains, one inserting an update on
a different slot after the queried rebinding and one reassigning the slots of all updates
that follow it. The remaining two readouts are constrained by the objective directly: a
paired difference between two valid training documents whose ground truths differ,
obtained by rewriting the queried slot's final value to a fresh one, and a single-document
margin between the superseded value and the truth, which is not a paired difference at all
and so cannot be moved by changing behaviour on an edited document alone. Collision rates with the ground truth (Table~\ref{tab:collide}) are 0.000 for
\textsc{primacy} in both cells, 0.115 and 0.003 for \textsc{frequency}, 0.161 and 0.353
for \textsc{position}, and 0.371 and 0.900 for \textsc{global last update}, the pairs being
the $R_{\mathrm{old}}=3$ and $R_{\mathrm{old}}=16$ cells. Eight readouts are measurable at
$R_{\mathrm{old}}=3$ and seven at $16$, where one edit's domain falls below our floor of 40
document pairs.

\paragraph{The window in which they can be compared.} Two conditions must hold at once. The
readout requires contrast-pair mass $\geq 0.5$, which no readout attains before the circuit
forms --- at $R_{\mathrm{old}}=3$ mass stays below 0.15 for all seven readouts that have a contrast
pair until step 5000, a property of the model's early flatness rather than of any edit,
since the multiplicity inversion's mass is as high as any control's there. And $g_L$ must be estimable, which we
require as $\cos(g_L^A, g_L^B) \geq 0.3$ on two halves of the loss batch; this fails from
step 12000 at $R_{\mathrm{old}}=3$ and step 8000 at $16$. The two windows overlap in ten
checkpoints, four in the first cell and six in the second, and we discard the one further
checkpoint that clears both but precedes the gate ($R_{\mathrm{old}}=16$, step 400, where
the readout is the plateau artifact of \S\ref{sec:plateau} whatever its mass). Those ten
are Table~\ref{tab:flatctrl}.

\begin{table}[ht]
\centering
\small
\caption{Eight readouts at the ten checkpoints where every readout with a contrast pair is valid
(mass $\geq 0.5$) and $g_L$ is estimable
($\mathrm{half} = \cos(g_L^A, g_L^B) \geq 0.3$). Columns \emph{alias},
\emph{t-diff} and \emph{margin} are $\lvert\cos(g_L, g_\Delta)\rvert$ for the
multiplicity inversion, for the paired difference whose two ground truths differ, and for
the single-document margin; chance is $2.0\times10^{-4}$ throughout and \emph{mass} is the
aliased readout's. The two ratio columns are t-diff and margin over alias. The five
remaining edits are omitted here and summarized in the text: their $\lvert\cos\rvert$ is
not ordered by collision rate at any checkpoint.}
\label{tab:flatctrl}
\begin{tabular}{rr rr rrr rr}
\toprule
& & & & \multicolumn{3}{c}{$\lvert\cos(g_L, g_\Delta)\rvert$}
& \multicolumn{2}{c}{ratio to alias} \\
\cmidrule(lr){5-7} \cmidrule(lr){8-9}
$R_{\mathrm{old}}$ & step & half & mass & alias & t-diff & margin
& t-diff & margin \\
\midrule
3  & 5000  & 0.599 & 1.00 & 0.00877 & 0.03965 & 0.03347 & $4.5\times$  & $3.8\times$ \\
3  & 6000  & 0.535 & 1.00 & 0.00317 & 0.02532 & 0.02199 & $8.0\times$  & $6.9\times$ \\
3  & 8000  & 0.350 & 1.00 & 0.01437 & 0.00317 & 0.00576 & $0.2\times$  & $0.4\times$ \\
3  & 10000 & 0.405 & 1.00 & 0.00060 & 0.00138 & 0.00058 & $2.3\times$  & $1.0\times$ \\
\addlinespace
16 & 1000  & 0.514 & 0.99 & 0.01838 & 0.04785 & 0.04113 & $2.6\times$  & $2.2\times$ \\
16 & 2000  & 0.474 & 0.95 & 0.00056 & 0.00428 & 0.00409 & $7.6\times$  & $7.3\times$ \\
16 & 3000  & 0.367 & 0.95 & 0.00683 & 0.00530 & 0.00135 & $0.8\times$  & $0.2\times$ \\
16 & 4000  & 0.344 & 0.89 & 0.00464 & 0.00510 & 0.00730 & $1.1\times$  & $1.6\times$ \\
16 & 5000  & 0.320 & 0.94 & 0.00012 & 0.00240 & 0.00513 & $20.5\times$ & $43.9\times$ \\
16 & 6000  & 0.334 & 0.86 & 0.00776 & 0.00985 & 0.01075 & $1.3\times$  & $1.4\times$ \\
\bottomrule
\end{tabular}
\end{table}

\paragraph{The aliased direction is flat.} Over the ten, removing the $g_L$ component
changes $\lVert g_\Delta \rVert$ by less than $1.7\times10^{-4}$; equivalently
$\lvert\cos(g_L, g_\Delta)\rvert$ is at most 0.018 against a chance level of
$1/\sqrt{2.6\times10^{7}} = 2.0\times10^{-4}$. These are one fact and not two, since the
rejected fraction is $\cos^2/2$ to first order. Curvature along $\hat u_{\Delta\perp}$ is
one to two orders of magnitude below curvature along $\hat u_L$.

\paragraph{So is every one of the seven.} Three readings, all against
Table~\ref{tab:flatctrl} and the per-checkpoint detail behind it. First, collision rate
does not order the edits. The rank correlation between collision rate and
$\lvert\cos\rvert$ within each checkpoint is $+0.203$, $+0.464$, $-0.145$, $-0.348$,
$-0.200$, $-0.100$, $-0.600$, $+0.700$, $-0.900$ and $+0.700$ --- four positive, six
negative, averaging $-0.023$ --- over the five or six edits available at each. Second, the
two constrained readouts do exceed the aliased one more often than not, in 8 and 7 of the
ten, but neither is significant on a one-sided binomial test ($p = 0.055$ and $p = 0.17$)
and the ratio ranges from $0.2\times$ to $44\times$. Its largest value comes from a
checkpoint where the aliased $\lvert\cos\rvert$ is $1.2\times10^{-4}$, below chance, so
that ratio is a division by noise. Third, $\lvert\cos\rvert$ is not stable even for one
fixed readout: for the aliased pair it spans $1.2\times10^{-4}$ to $1.8\times10^{-2}$
across the ten, a factor of 153, with the three smallest at $0.6$, $2.9$ and $3.1$ times
chance, the first of them below it. Curvature does not order the readouts either: at $R_{\mathrm{old}}=3$, step 5000
the aliased direction is near the lowest of the eight and the highest belongs to a rule
colliding at 0.371, while at $R_{\mathrm{old}}=16$, step 5000 it is second highest of the seven.

\paragraph{Why the test fails.} Two reasons, both structural rather than particular to this
implementation. A unit vector in $2.6\times10^{7}$ dimensions obtained by rejecting $g_L$
from any gradient has curvature set by the bulk of the loss surface rather than by the
readout that produced it: across the eight readouts at
$R_{\mathrm{old}}=3$, step 16000 --- a checkpoint outside the ten, where the quadratic
relation does not require $g_L$ to be estimable --- the induced $\Delta L$ equals the
quadratic term
$\tfrac12 \varepsilon^2 \hat u^\top H_L \hat u$ to within 15\%, so $\Delta L$ and the
curvature are one measurement and neither is a property of the readout. Separately, the
answer is one token of roughly two hundred and its cross-entropy is saturated once the
circuit forms, so any readout defined at that position is nearly orthogonal to the
token-averaged loss gradient whatever it measures. The second reason worsens with
training: $\cos(g_L^A, g_L^B)$ falls from 0.82 and 0.64 at step 400 to between 0.03 and
0.13 at 14000--16000 steps, so at the budget where the paper's readouts are taken $g_L$ is
itself mostly noise and orthogonality to it carries little information. The two conditions
of the previous paragraph are in tension for the same reason: mass requires late, $g_L$
requires early.

\paragraph{What we conclude.} The direction that trades \textsc{recency} against
\textsc{rarity} is flat relative to what the objective uses. So is every other direction we
constructed the same way, including two the objective constrains. We therefore rest no
claim on this measurement. \S\ref{sec:flat} rests on Equation~\ref{eq:alias}, an identity
holding on every document (Table~\ref{tab:collide}), together with the variance ordering
it reports; the geometry was to have been a third and independent line of evidence and
it is not one. The negative result is itself bounded: it covers two cells in one seed, at
ten checkpoints, with the aliased readout's $\lvert\cos\rvert$ falling to chance at one of
them.

\end{document}